\documentclass[letter paper,10pt]{article} 

\newcommand\authormark[1]{\textsuperscript{#1}}

\usepackage{amsmath}
\usepackage[a4paper, total={6in, 9.3in}]
{geometry}
 \usepackage{graphicx}
\usepackage[colorlinks=true, bookmarks=false,citecolor=blue,urlcolor=blue]{hyperref}
\usepackage{cleveref}
\usepackage{xcolor}

\usepackage{fancyhdr}
\fancypagestyle{default}{
  \fancyhf{}
  \fancyfoot[R]{\thepage}
  \renewcommand{\headrulewidth}{0pt}
}

\begin{document}


\date{}
\title{Achieving more human brain-like vision via human EEG representational alignment}
\author{}
\maketitle
\vspace{-1.5cm}
\begin{center}
    Zitong Lu \authormark{1,2,*}, Yile Wang \authormark{3}, Julie D. Golomb\authormark{1}
\end{center}

\begin{center}
$^{1}${Department of Psychology, The Ohio State University}\\
$^{2}${McGovern Institute for Brain Research, Massachusetts Institute of Technology}\\
$^{3}${Department of Neuroscience, The University of Texas at Dallas}\\
\end{center}

\begin{center}
    *email: zitonglu@mit.edu
\end{center}
\pagestyle{default}
\renewcommand\headrulewidth{0pt}
\vspace{0.5cm}

\vspace{40pt}

\section*{Abstract}

Despite advancements in artificial intelligence, object recognition models still lag behind in emulating visual information processing in human brains. Recent studies have highlighted the potential of using neural data to mimic brain processing; however, these often rely on invasive neural recordings from non-human subjects, leaving a critical gap in understanding human visual perception. Addressing this gap, we present, ‘Re(presentational)Al(ignment)net’, a vision model aligned with human brain activity based on non-invasive EEG, demonstrating a significantly higher similarity to human brain representations. Our innovative image-to-brain multi-layer encoding framework advances human neural alignment by optimizing multiple model layers and enabling the model to efficiently learn and mimic the human brain’s visual representational patterns across object categories and different modalities. Our findings suggest that ReAlnets better align artificial neural networks with human brain representations, making it more similar to human brain processing than traditional computer vision models, which takes an important step toward bridging the gap between artificial and human vision and achieving more brain-like artificial intelligence systems.

\vspace{20pt}

\noindent\textbf{Key words}: Object Recognition; Neural Alignment; Human Brain-Like Model

\newpage

\section{Introduction}

While current vision models in artificial intelligence (AI) have achieved remarkable advancements, they still fall short of capturing the full complexity and adaptability of the human brain's information processing. Deep convolutional neural networks (DCNNs) now rival human performance in object recognition \cite{Lecun2015}, and many studies have identified representational similarities between the hierarchical structures of DCNNs and the ventral visual stream \cite{Cichy2016, Guclu2015, Kietzmann2019, Yamins2014, Lu2023c}. However, aligning DCNNs with human neural representations, though promising, remains an area with significant potential for further exploration. Enhancing the similarity between visual models and the human brain has become a critical concern for both computer scientists and neuroscientists. From a computer vision perspective, brain-inspired models often exhibit greater robustness and generalization, which are essential for achieving brain-like intelligence. Meanwhile, from a cognitive neuroscience perspective, models that more closely resemble brain representations can provide valuable insights into the mechanisms of human visual processing. 

Conventional approaches - increasing model depth and layer count - have struggled to emulate the complexity of human visual processing \cite{Rajalingham2018}. Researchers have proposed various bio-inspired strategies to leverage the understanding of the human brain to enhance current AI vision models, including \textit{ altering the architecture of the model} (adding recurrent structures \cite{Kar2019, Kietzmann2019, Kubilius2019, Spoerer2017, Tang2018}, dual-pathway models \cite{Bai2017, Choi2023, Han2022, Han2023, Sun2017}, topographic constraints \cite{Finzi2022, lee2020, lu2023e, Margalit2023} or feedback pathways \cite{Konkle2023} ) and \textit{changing the training task} (using self-supervised training \cite{Konkle2022, Prince2023} or 3D task models \cite{OConnell2023}). However, an important question remains: \textbf{Can we directly use human neural activity to align ANNs in object recognition, thereby achieving more human brain-like vision models?}

Several previous studies have begun to explore the integration of neural data into machine learning, particularly deep learning models, to enable these models to learn biologically inspired representations from neural data. One common strategy is to introduce a similarity loss during training to increase the representational alignment between models and neural activity, often derived from mouse V1 or monkey V1 and IT regions  \cite{Dapello2023, Federer2020, Li2019, Pirlot2022}. Another strategy from \cite{Safarani2021} integrates an additional task that uses an encoding module to predict monkey V1 neural activity. Both similarity-based methods and multi-task frameworks have been shown to produce more brain-like representations and improve model robustness. However, the key challenge of these neural alignment studies is they can only do single brain region or single model layer alignment. Previous studies typically focused on aligning only a single layer of a CNN with a specific brain region, such as V1 or IT, without a clear understanding of how multiple layers correspond to different brain regions. This oversimplification can lead to misalignment and inaccuracies. Moreover, most of these studies depend on invasive neural recordings from animals instead of human neural data, which limits the direct applicability of findings to human visual processing. Human non-invasive recordings, such as fMRI and EEG, often have lower data quality compared to invasive animal recordings, making it more challenging for models to learn human brain representations effectively. Early attempts involved applying human fMRI signals as additional inputs to machine learning classifiers such as SVMs and CNNs, resulting in improved category classification performance without altering the internal feature representations of the models themselves \cite{Fong2018, Spampinato2017, Palazzo2021}. While more recent research has directly optimized CNN internal representations align with human fMRI data for video emotion recognition \cite{Fu2023}, going beyond earlier approaches that only incorporated fMRI features into classification tasks \cite{Fong2018}, this approach has been limited to a relatively simple six-category emotion classification task. It remains unclear whether it could scale effectively to more complex domains like object recognition, which involves a far greater diversity of categories (e.g., 1000 classes in ImageNet-trained models). Also, it is unclear whether we can apply human neural data to optimize ANNs to achieve more human brain-like internal model representations.

To address these limitations, our study proposes a novel approach that employs an encoding-based framework for multi-layer alignment. This framework goes beyond simple similarity by training an additional encoding module to predict human neural activity, thereby enabling the model to autonomously extract complex visual features. Our approach leverages human neural data to achieve more effective alignment with human brain representations in object recognition tasks.

To bridge the gap between AI vision and human vision, we introduce Re(presentational)Al(ignment)net framework (hereafter, the ReAlnet framework), a method for effectively aligning vision models with human brain representations obtained from non-invasive EEG recordings. EEG was chosen for its high temporal resolution and the capacity to collect a large number of trials through rapid successive stimulus presentation, making it a cost-effective and scalable option. Our novel encoding-based multi-layer alignment framework effectively allows neural networks to learn human brain representations, enabling the creation of personalized vision models tailored to individual neural data. In this ReAlnet framework, each individualized ReAlnet refers to one EEG-aligned model instance trained on a single subject's data, while the plural ReAlnets refers to the set of ten such individualized models. Here, we define "direct alignment" as modifying a model’s internal representational structure via a learning objective that explicitly incorporates human neural data. Under this definition, our approach constitutes the first direct alignment of object recognition models using non-invasive human EEG signals. This novel approach opens new possibilities for enhancing brain-like representations in AI models. Furthermore, human EEG-optimized ReAlnets demonstrate improved alignment with human brain representations across different modalities (both human EEG and fMRI) and human behaviors.

\vspace{20pt}

\begin{figure}[ht!]
\vskip 0in
\begin{center}
\centerline{\includegraphics[width=\columnwidth]{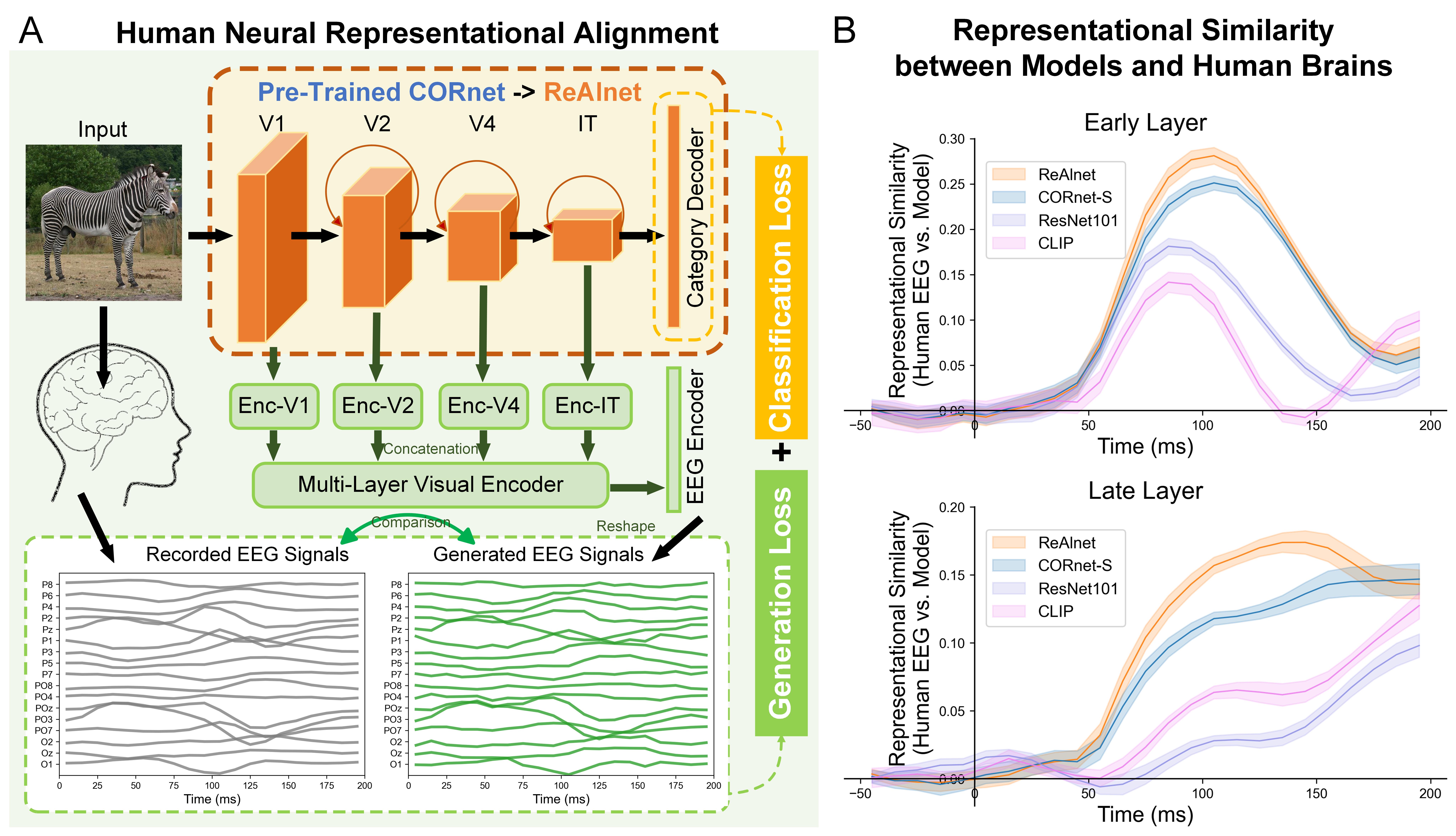}}
\caption{ReAlnets aligned with human EEG signals as more human brain-like vision models. (A) An overview of ReAlnet alignment framework. Adding an additional multi-layer encoding module to an ImageNet pre-trained CORnet-S, the outputs contain the category classification results and the generated EEG signals. Using THINGS EEG2 training dataset, we aim to minimize both \textit{classification loss} and \textit{generation loss}, enabling CORnet to not only stabilize the classification performance but also effectively learn human brain features and transform into ReAlnets. (B) Representational similarity between internal representations in models and human temporal EEG signals from THINGS EEG2 test dataset. Models include ReAlnets and their primary comparison model CORnet-S, along with ResNet101 and CLIP (with a ResNet101 backbone) as additional baselines. Because the additional baselines have different number of layers, for all models we took the first layer as the Early Layer, and the layer before the classification layer (or last visual layer in CLIP) as the Late Layer for this analysis. The line labeled "ReAlnet" reflects the mean similarity across 10 individual ReAlnets, each trained on a different subject's EEG data (N=10). For comparison models, each line reflects the mean similarity between the same 10 human EEG datasets and the single model instance. ReAlnets} consistently show the highest similarity to the human brain. Lines and shading reflect mean±SEM.
\label{Figure1}
\end{center}
\vskip -0.2in
\end{figure}

\section{Results}

\subsection{Aligning CORnet with human EEG representations}

In this study, we build and test a novel image-to-brain multi-layer encoding alignment framework (\Cref{Figure1}). Our core aim is to investigate whether aligning the model with individual neural representations of humans can enhance the model’s similarity to the human brain. As an intial fidelity check, we confirmed that the approach of jointly minimizing classfication and generation losses during the training (\Cref{Figure1}A, see Methods) yields accurate classification of object category (average Top-1 and Top-5 accuracy on ImageNet: 69.31\% and 88.88\%, \Cref{FigureS1}A-B) and generates realistic EEG signals (77.85\% similarity, computed as Spearman correlation between generated and real EEG time series, treating the full channel × timepoint matrix of each image as a vector, averaged across the test set, \Cref{FigureS1}C-D). While the latter demonstrates that the model successfully learned to reproduce realistic EEG activity, building the most accurate EEG generator is not our main objective. Instead, we seek to use EEG generation as a training mechanism to optimize the model's \textit{internal representations}, with the goal of making them more similar to human brain representations. Our critical hypthesis is that modifying the model's internal representations in this way may carry benfits that generalize across image diet and modalities. In the following sections, we evaluate whether the model's internal representations have indeed become more brain-like, by conducting representational similarity analysis (RSA to test models' similarity to human EEG, human fMRI, and behavior.

To apply this alignment framework, we built ten individual ReAlnets, using the state-of-the-art CORnet-S model \cite{Kubilius2018, Kubilius2019} as the foundational architecture. Each ReAlnet, which has the same architecture as CORnet, is additionally trained on a real human subject’s EEG signals, recorded while viewing a massive number of natural images from THINGS EEG2 \cite{Gifford2022} training set. After training, we employed an independent test dataset consisting of 200 images and associated EEG activity from the THINGS EEG2 test set. These test set images had not been presented at all during the training process, coming from entirely novel (untrained) object categories. We input these 200 test images to each model and obtained the feature vectors corresponding to each image for each layer in the model. Then we calculated the temporal similarity between different models and human brain EEG using RSA method. {\Cref{Figure1}B} shows the representational similarity between different models and temporal human brain EEG signals for early and late layers. Comparing the ReAlnets to the original CORnet-S, along with ResNet101 and CLIP (with a ResNet101 backbone) as additional baselines, ReAlnets consistently show the highest similarity to the human brain patterns.

Our assessment of the models' similarity to humans is not limited to its similarity with human brain EEG representations (based on THINGS EEG2 test set \cite{Gifford2022}); Similar to EEG, we then evaluated the model's similarity to human brain fMRI representations (a completely different modality) from human subjects viewing novel image categories (based on Shen fMRI test set \cite{Shen2019}). Additionally, we measured the similarity between the model and human behavior in several object recognition tasks using the Brain-Score platform \cite{Schrimpf2020} based on two behavioral benchmarks. For methodological details, refer to our Methods section. In the following sections, we elaborate on how ReAlnets -- compared with the original CORnet not aligned with human neural data -- more closely resembles human brain representations.

\subsection{Improved similarity to human EEG}

Here, for each of the 10 human subjects from the THINGS EEG2 dataset \cite{Gifford2022}, we calculated (1) the similarity between their EEG data and CORnet (with the same structure as ReAlnets, but non aligned human neural data and non-individualized model), and (2) the similarity between their EEG data and the subject-matched ReAlnet, via RSA based on THINGS EEG2 test dataset (See more details in Methods section). As an additional control, we trained a Scrambled-model version of the ReAlnet framework by aligning on 10 subjects' scrambled EEG signals (more details in Control Experiments section and Methods section) and conducted the same RSA analysis. 

ReAlnets exhibit significantly higher similarity to human EEG neural dynamics for all four visual layers (Layer V1: 70-130ms and 160-200ms; Layer V2: 60-200ms; Layer V4: 60-200ms; Layer IT: 70-160ms) than the original CORnet without human neural alignment, or than Scrambled models trained on scrambled EEG signals (\Cref{Figure2}A). The EEG similarity curves often show a two-peaked shape, potentially reflecting different stages of visual information processing. Based on prior studies on human neural dynamic representations \cite{Lu2023d, Teichmann2024, Khaligh-Razavi2018}, the earlier peak (\~100ms) likely corresponds to the processing of lower-level visual features, such as color and retinal size. And the later peak (\~150-200ms or even later) may reflect the processing of higher-level semantic features, such as real-world size and animacy information.

\begin{figure}[ht!]
\vskip 0in
\begin{center}
\centerline{\includegraphics[width=0.9\columnwidth]{./Figure2.jpg}}
\caption{ReAlnets show higher similarity to human EEG and hierarchical individual variability. (A) Representational similarity time courses between human EEG and models (ReAlnets, Scrambled models, and CORnet) for different layers respectively. Dark blue square dots at the bottom indicate the timepoints where ReAlnet vs. CORnet were significantly different ($p<.05$). Grey square dots at the bottom indicate the timepoints where ReAlnet vs. CORnet were significantly different ($p<.05$). Lines and shading reflect mean±SEM. (B) Similarity improvement and similarity improvement ratio of ReAlnets compared to CORnet at the similarity peak timepoint. Each circle dot indicates an individual ReAlnet. Error bar reflects ±SEM. (C) Time courses of the maximum representational similarity between human EEG and different models (ReAlnets, Scrambled models, and CORnet), computed by taking the highest similarity across all model layers at each timepoint. Dark blue square dots at the bottom indicate timepoints where ReAlnets significantly outperformed CORnet ($p < .05$). Grey square dots indicate significant differences between ReAlnets and Scrambled models ($p < .05$). Lines and shading reflect mean±SEM. (D) Top: ReAlnet individual variability matrices of four visual layers. Bottom left: ReAlnet individual variability along layers. Bottom right: Human fMRI individual variability along the visual cortex. Each circle dot indicates a pair of two personalized ReAlnets or two human subjects. Error bar reflects ±SEM. (E) Cross-subject similarity matrix showing each individualized ReAlnet (rows) generalizes to EEG representations from all 10 subjects (columns). Each cell reflects the average representational similarity between human EEG and ReAlnets and CORnet across four model layers and the 50-200 ms time window. (F) Cross-subject generalization beyond baseline CORnet. Each cell reflects the ReAlnet–CORnet difference in EEG similarity, with positive values indicating that even mismatched ReAlnets outperform CORnet on other subjects’ EEG data. (G) Left: Column-wise normalized similarity matrix, where each column is scaled such that the highest similarity value is 1. Right: A statistical comparison between matched and mismatched pairs. Black asterisks indicate significantly higher similarity of matched pairs than mismatched pairs ($p<.05$). Error bar reflects ±SEM.}
\label{Figure2}
\end{center}
\vskip -0.2in
\end{figure}

Further statistical analysis of each layer’s similarity improvement (ReAlnet - CORnet) and improvement ratio ((ReAlnet - CORnet) / CORnet) also indicates that at the similarity peak timepoint, there is a maximum of a 6\% similarity improvement and a 40\% improvement ratio (\Cref{Figure2}B). Importantly, all indiviudalized ReAlnets showed positive improvement over CORnet, suggesting that our ReAlnet framework robustly generalizes across subjects that effectively enhances model-EEG similarity using each individual's EEG data. It is worth noting that the test set doesn't overlap with the training set in terms of object categories (concepts). Therefore, these significant improvements reveal ReAlnets' generalization capability across different object categories. In addition to report layer-wise RSA curves, we also computed the maximum similarity across all layers for each model at each time point to provide a clearer summary of model-EEG alignment (\Cref{Figure2}C).

These results suggest three findings: (1) Our multi-layer alignment framework indeed improves all layers’ similarity to human EEG representations. (2) Every ReAlnet with individual neural alignment exhibits improved similarity to human EEG compared to the basic CORnet. (3) ReAlnets demonstrate the generalization of improvement in human brain-like similarity across object categories, as the image categories used for testing were entirely absent during the alignment training.

Additionally, unlike traditional models in computer vision, ReAlnet is a personalized model trained based on different individuals' neural data. Just as neuroscientists are interested in studying individual differences in the brain, we are similarly interested in exploring individual differences across the ten personalized ReAlnets - whether ReAlnets exhibit intra-model individual variabilities and how such variabilities change across different layers of the model. Note that CORnet serves as a single, publicly available reference model with fixed official weights, and thus does not have individualized versions analogous to ReAlnets. Therefore, inter-model variability can only be meaningfully assessed across individualized ReAlnets. Importantly, all ten ReAlnets were initialized with the same pretrained CORnet-S weights and trained using the same random seed, ensuring that any observed differences arise solely from the individual EEG data used for fine-tuning, not from variations in model initialization. We hypothesize that the observed individual differences in ReAlnets may provide insights into potential mechanisms underlying individual differences in human brains. To investigate this, we also analyzed individual variability in human brain regions (V1, V2, V4, and LOC) using the Shen fMRI dataset based on a similar RDM-based correlation analysis. As a comparison, we also conducted comparisons between model RDMs based on 200 images in THINGS EEG2 test set across different layers, using the dissimilarity (one minus the Spearman correlation coefficient) between two RDMs corresponding to two ReAlnets as an individual variability index. Importantly, we included the output layer of ReAlnets, which performs category classification, and found that its individual variability decreases, mirroring the pattern seen in the LOC region of the human brain (additional analysis in \Cref{FigureS2} confirms that LOC is significantly more similar to the model's output layer than Layer IT, suggesting that the model's output layer rather than Layer IT is more comparable to brain LOC region). While humans also show hierarchical increases in between-subject variability, the larger absolute scale may partly arise from cross-individual measurement noise in neuroimaging. Our results suggest: (1) Personalized ReAlnets indeed exhibit individual variability (\Cref{Figure2}D). (2) This variability increases with the depth of the layers (from Layer V1 to Layer IT, \Cref{Figure2}D) and decreases in the output layer, which aligns with the trend observed in the human brain - where individual variability increases from V1 to V4 and then decreases in LOC.

To further evaluate the generalization ability of individualized ReAlnets within the EEG modality, we conducted a cross-subject analysis, testing each model against the EEG RDMs of all subjects. We computed a model–subject similarity matrix where each entry represents the average RSA similarity across four model layers and the 50-200 ms time window. This approach summarizes each ReAlnet’s representational match with every subject and produces a compact confusion matrix (\Cref{Figure2}E). To more directly assess cross-subject generalization, we subtracted each subject’s similarity with CORnet from all cells in that subject’s column (\Cref{Figure2}F). Positive values in this normalized matrix indicate that even mismatched ReAlnets outperform the CORnet baseline, confirming that ReAlnets capture representational structures that generalize across individuals. Finally, to control for differences in signal quality across subjects, we applied column-wise normalization to the matrix (\Cref{Figure2}G). To more directly quantify subject-specificity, we compared similarity values for matched model-subject pairs (the diagonal: i.e., models evaluated on the same subject used for training) against unmatched pairs (the off-diagonal) and observed that matched model–subject pairs show significantly higher similarity than unmatched pairs ($t$=5.6068, $p$=.0003). These results confirm that individualized ReAlnets capture subject-specific representational features, and that these models generalize across individuals within the same neural modality.

\subsection{Improved similarity in ReAlnets to human fMRI}

Although ReAlnets demonstrate higher similarity to human EEG, a question arises: Do ReAlnets learn representations specific to EEG, or more general neural representations of the human brain? To ensure that our alignment framework enables the model to learn representations beyond the single modality of EEG, we utilized additional human fMRI data of three human subjects viewing natural images to evaluate the model’s cross-modality representational similarity to human fMRI.

\begin{figure}[ht!]
\vskip 0in
\begin{center}
\centerline{\includegraphics[width=\columnwidth]{./Figure3.jpg}}
\caption{ReAlnets show higher similarity to human fMRI representations. Representational similarity between models and human fMRI of five different brain regions when three subjects in Shen fMRI test dataset viewed (A) natural images, (B) artificial shape images, and (C) alphabetical letter images. Black asterisks indicate significantly higher similarity of ReAlnets than that of Scrambled model or CORnet ($p<.05$), and grey asterisks indicate significantly lower similarity of ReAlnets than that of Scrambled model or CORnet ($p<.05$). Each circle dot indicates an individual ReAlnet or Scrambled model. Error bar reflects ±SEM.}
\label{Figure3}
\end{center}
\vskip -0.2in
\end{figure}

Excitingly, we indeed observed a clear increase in this cross-modal brain-like similarity. Based on human fMRI signals of three subjects viewing 50 natural images, the similarity results indicate that, overall, ReAlnets exhibit a stronger resemblance to human fMRI data than CORnet and Scrambled models, despite being aligned using a different type of neural signal with very different spatiotemporal properties (human EEG data), and that was collected from a different set of participants (\Cref{Figure3}A). Although a few conditions show CORnet may have higher similarity than ReAlnets, the general trend favors ReAlnets - highlighting their stronger alignment with human neural representations and showing that EEG-optimized ReAlnets generalize broadly to human representations. Moreover, the similarity improvements across modalities were significantly correlated (\Cref{FigureS3}): the ReAlnets showing more similarity improvement when tested on EEG data also showed more similarity improvement when tested on fMRI data, ($r=.9204, p<.001$), indicating that the ReAlnet alignment framework is tapping into meaningful and robust human neural signals.

Additionally, to further assess the enhanced similarity between ReAlnets and human brain representations during image observation - not solely with natural images - we also measure the similarity between models and human fMRI signals from three subjects viewing 40 artificial shape and 10 alphabetical letter images in the Shen fMRI test dataset. Although these images might be outside the natural image distribution, our results further demonstrate ReAlnets' improved similarity to human brain representations in comparisons to CORnet and Scrambled models (\Cref{Figure3}B-C). While some comparisons yield small absolute differences, we also examined effect sizes to better assess their statistical and practical significance (detailed statistical results including $t$-value, $p$-value, and Cohen's $d$ are listed in \Cref{TableS1}). The stronger effects ($p<.001$ and |Cohen's $d$|>1.0; bolded in \Cref{TableS1}) were observed in brain region V2 and V4 under natural images condition and in brain region V1 and V2 under artifical shape images condition, indicating particularly robust representational improvements in these regions under different conditions. Other ROIs and conditions also showed statistically significant improvements, though with relatively smaller effect sizes, suggesting that the enhancement is widespread but especially pronounced in certain areas.

These findings further highlight three points: (1) Across multiple ROIs, ReAlnets exhibits higher human fMRI similarity than CORnet. (2) Despite being trained with the EEG data of subjects not in the fMRI dataset, almost every ReAlnet shows higher fMRI similarity, suggesting that ReAlnets learn consistent brain information processing patterns across subjects. (3) Images from the fMRI datasets for evaluation were never presented during the alignment training, reaffirming the generalization of ReAlnets in improving brain-like similarity across object categories and images.

\subsection{Improved similarity in ReAlnets to behavior}

\begin{figure}[ht!]
\vskip 0in
\begin{center}
\centerline{\includegraphics[width=0.8\columnwidth]{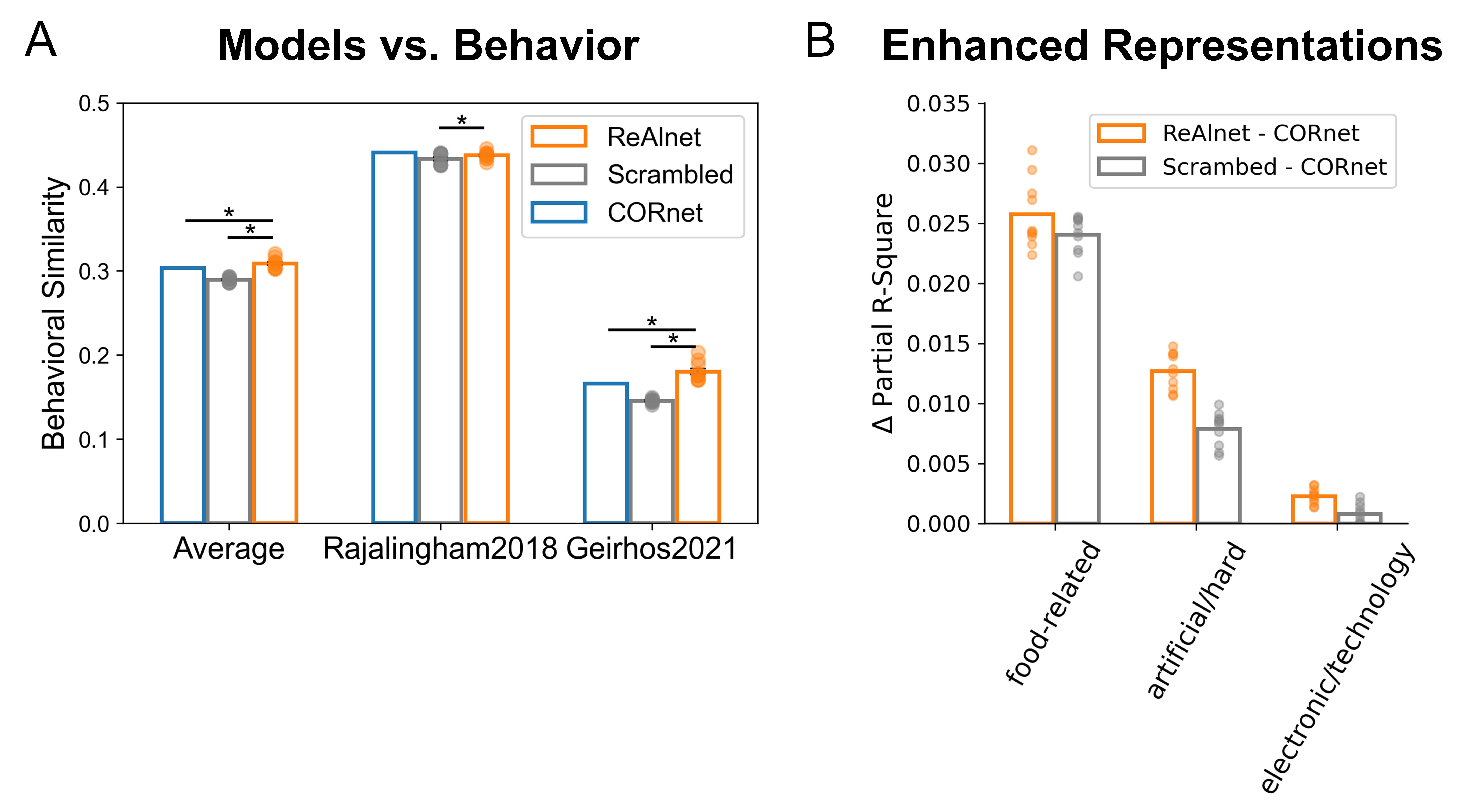}}
\caption{Enhanced behavioral similarity and feature representations in ReAlnets. (A) ReAlnets show higher similarity to human behavior based on the Brain-Score platform. Each orange circle dot indicates an individual ReAlnet. Each grey circle dot indicates an individual scrambled model. Asterisks indicate significantly higher similarity of ReAlnets than that of CORnet or scrambled models ($p<.05$). (B) Top-3 enhanced feature representations in ReAlnets compared to CORnet and scrambled models. Each orange circle dot indicates an individual ReAlnet. Each grey circle dot indicates an individual scrambled model. Error bar reflects ±SEM.}
\label{Figure4}
\end{center}
\vskip -0.2in
\end{figure}

Does this neural-level alignment also translate into any behavioral alignment? To test whether ReAlnets show improved similarity to human behavior, we calculated the scores of CORnet and 10 personalized ReAlnets based on the human behavioral assessments, including two object recognition tasks, in the Brain-Score platform \cite{Schrimpf2020}. One task compared how well the ANN, compared to primates and humans, could recognize objects presented in the center of their visual field, even when the objects varied in position, size, viewing angle, and background \cite{Rajalingham2018}. The other paradigm tested the similarity of behavioral error between the errors made by humans and ANN on an image-by-image basis \cite{Geirhos2021}. These scores serve as indicators of the models' similarity to human behavior. The average of two behavioral scores were used to compare model-behavior similarity. Excitingly, the result reveals that ReAlnets, aligned with human EEG data, exhibit representations significantly more akin to human behavior than CORnet does (ReAlnets vs. CORnet: $t=2.7702, p=.0217, d=.8762$; ReAlnets vs. Scrambled models: $t=8.5582, p<.0001, d=3.8273$) (\Cref{Figure4}A), further expanding and emphasizing ReAlnets' status as a more human brain-like vision model. Especially, the results show that ReAlnets exhibit a greater improvement in human behavior similarity on the Geirhos2021 task compared to Rajalingham2018 task. However, the Rajalingham2018 task uses grayscale images where objects are manipulated by varying position, size, viewing angle, and background, while the Geirhos2021 task employs out-of-distribution colorful natural images. A possible explanation is that because ReAlnets are trained with EEG data recorded from subjects viewing natural object images, it better captures human-like behavioral consistency when evaluated on naturalistic, colorful stimuli. Of course, the differences in image manipulation between the two tasks may also contribute to this disparity.
 
In addition, we submitted our ReAlnets to the updated Brain-Score platform for evaluation and made the scores publicly available on the Brain-Score website (\href{https://www.brain-score.org/vision/}{https://www.brain-score.org/vision/}). Consistently, the results confirm that ReAlnets achieve significantly higher scores compared to CORnet (see \Cref{FigureS4}), further supporting their improved alignment with brain visual processing.

\subsection{Refined object feature representations in ReAlnets}

We found that ReAlnets exhibit improved similarity to the human visual system, suggesting that by incorporating human EEG data, they have learned brain representations that purely image-trained models could not capture. Thus, this raises an important question: how do their internal representations differ from those of CORnet? To further investigate what visual object feature representations have been enhanced in ReAlnets by learning human brain signals, we conducted internal representational analysis on purely image-trained CORnet and human EEG-aligned ReAlnets (For the detailed description of the methodology used to analyze these enhanced feature representations, refer to our Methods - Model internal representational analysis section). Here, we tracked feature representations of 49 object feature dimensions from THINGS (\cite{Hebart2020}) on models' layer IT to see which features that could be better captured by ReAlnets than CORnet. These 49 object feature dimensions, derived from the THINGS dataset, describe various conceptual and perceptual properties of objects, providing a structured way to assess how different models encode object features beyond simple category labels. \Cref{Figure4}B shows that ReAlnets show significantly enhanced representations in food-related, artificial, and electronic information (more detailed results of all 49 object feature dimensions are shown in \Cref{FigureS5}). This result suggests that although CORnet trained solely on image data can capture many higher-level object features, some object features, such as food-related information, showed enhanced representation in ReAlnets compared to CORnet. Interestingly, Scrambled models also showed enhanced representations for some of these features, suggesting that even the statistical properties of EEG data can refine internal model representations — though to a lesser extent than real, temporally intact EEG signals.

One might wonder whether these refined object feature representations were simply due to a change in visual "diet" - that is, the distribution of stimulus categories - when switching from ImageNet to THINGS EEG2 images during alignment. To directly address this potential confound, we conducted a control experiment in which we removed all food-related image-EEG pairs from the training set in THINGS EEG2 prior to the training process. Specifically, we excluded all training images with a "food" label in THINGS dataset. We then retrained ReAlnets on the reduced dataset using the same alignment framework and evaluated their internal feature representations. Remarkably, even without any exposure to food-related training images or neural signals during our alignment training process, the food-depleted ReAlnets still exhibited significantly enhanced representation of food-related dimensions compared to the original CORnet (\Cref{FigureS6}). While the strength of this enhancement was somewhat reduced, the persistence of food-related representation — despite the complete removal of food stimuli — strongly supports the conclusion that the representational changes observed in ReAlnets are driven primarily by neural alignment rather than by image category frequency.

\subsection{Control experiments}

To systematically evaluate how different training manipulations influence model-to-brain alignment, we conducted a set of control experiments. On the one hand, we would like to test the importance of the two loss components in generation loss, reconstruction and contrastive learning losses, in the model alignment. On the other hand, prior non-EEG studies have shown that even optimizing CNNs with shuffled monkey electrophysiology or fMRI signals does not necessarily degrade model performance \cite{Federer2020,Shao2024}. Motivated by these, we conducted control experiments to test four aspects: (1) How does the reconstruction loss influence model-to-brain alignment? (2) How does contrastive learning loss influence model-to-brain alignment? (3) If we disrupt the pairing of each image with the EEG signal from the same subject but elicited by viewing a different image, or (4) If we scrambled the EEG time-series data itself, can the model still learn the neural representation patterns of the human brain? Accordingly, we trained four additional sets of ReAlnets based on human EEG data from ten subjects in THINGS EEG2 dataset, termed as \textit{W/o ContLoss} models (without the constrastive loss component), \textit{W/o MSELoss} models (without the MSE loss component), \textit{Unpaired} models (where the pairing between images and EEG signals was disrupted), and \textit{Scrambled} models (where the EEG time-series were scrambled). More detailed statistical results including t-value, p-value, and Cohen's d are listed in \Cref{TableS2,TableS3}.

\begin{figure}[h!]
\begin{center}
\centerline{\includegraphics[width=0.9\columnwidth]{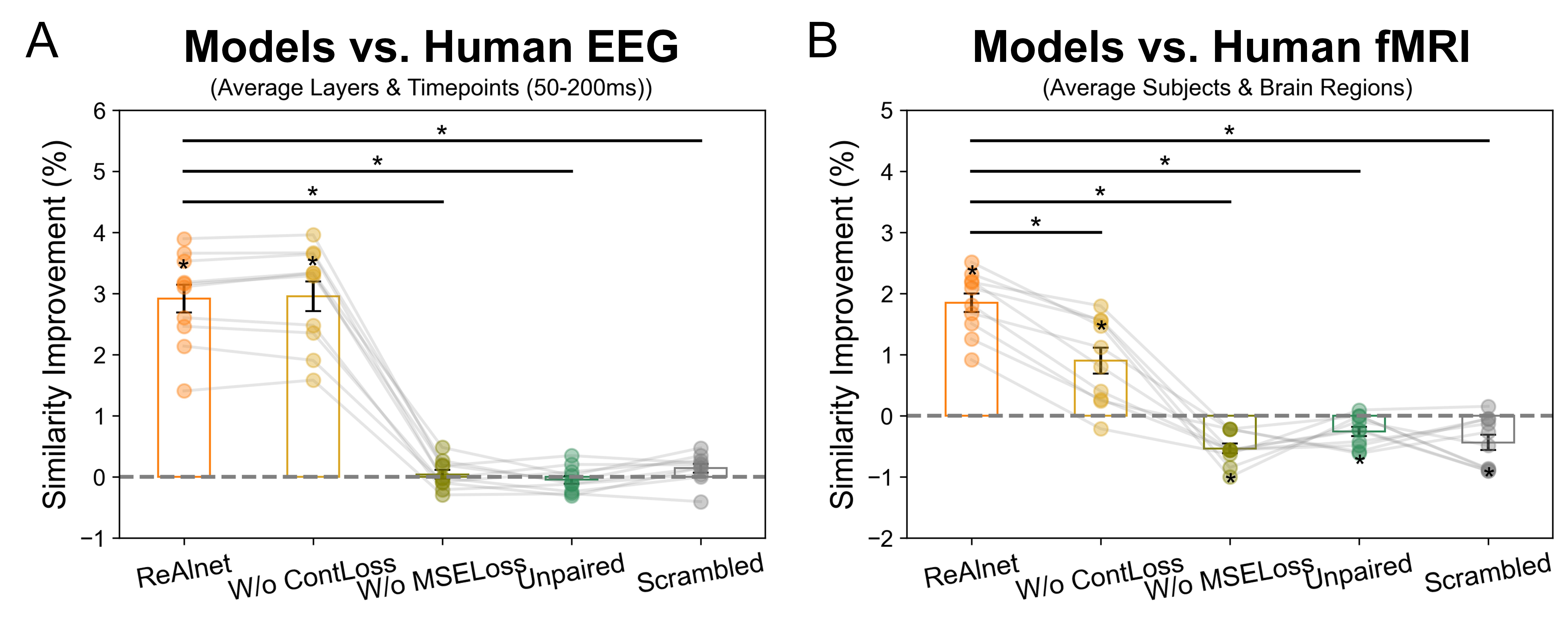}}
\caption{Results of control experiments. (A) Improvement in human EEG similarity of ReAlnets and control models compared to CORnet. (B) Improvement in human fMRI similarity of ReAlnets and control models compared to CORnet. Each circle dot indicates an individual model. Asterisks indicate the significance ($p<.05$). Error bar reflects ±SEM.}
\label{Figure5}
\end{center}
\end{figure}

We tested the control models on the THINGS EEG2 test dataset and the Shen fMRI dataset, and calculated the similarity improvement for each control model compared to ReAlnets and the CORnet baseline. \Cref{Figure5} plots the improvement in similarity (compared to CORnet baseline) for ReAlnets and the four controls. Here, we averaged the EEG similarity improvement of all layers and timepoints between 50 and 200ms, and averaged the fMRI similarity improvement of three subjects and five brain regions (See more detailed EEG and fMRI similarity results in \Cref{FigureS7}). The \textit{W/o MSELoss}, \textit{Unpaired}, and \textit{Scrambled} control models showed no significant similarity improvement over CORnet for human EEG (\textit{W/o MSELoss}: $t=.5640, p=.5865, d=.1784$; \textit{Unpaired}: $t=-.6654, p=.5225, d=-2104$; \textit{Scrambled}: $t=1.9450, p=.0835, d=.6154$) and even significant similarity decreasing for human fMRI (\textit{W/o MSELoss}: $t=-6.3837, p=.0001, d=-2.0187$; \textit{Unpaired}: $t=-3.1304, p=.0121, d=-.0900$; \textit{Scrambled}: $t=-3.2647, p=.0098, d=-1.0324$). The \textit{W/o ContLoss} control models were significantly improved over CORnet for both modalities (EEG: $t=11.4987, p<.0001, d=3.6362$; fMRI: $t=4.1036, p=.0027, d=1.2977$), but they didn't perform as well as ReAlnets in terms of fMRI similarity ($t=-8.0364, p<.0001, d=-1.5494$) but showed similar improvement for model-EEG similarity ($t=.8543, p=.4151, d=.0517$).

The results of the control experiments reveal: (1) \textit{W/o ContLoss} models still exhibit an improvement in human brain similarity compared to CORnet. However, while the similarity to human EEG did not decrease compared to ReAlnets, the similarity to cross-modality human fMRI significantly decreased. This suggests that the contrastive loss in our alignment framework enables ReAlnets to learn broader and more generalized visual representation patterns across different neuroimaging modalities, such as EEG and fMRI data. (2) The \textit{W/o MSELoss}, \textit{Unpaired}, and \textit{Scrambled} models failed to enhance brain similarity, which show no significant improvement in brain similarity compared to CORnet, indicating that the training process requires the model to effectively learn the specific neural visual features from the actual EEG signals corresponding to each image. Only in this way can the model become more human brain-like and then exhibit higher similarity to the human brain across different object images, categories, and human neuroimaging data modalities.

\subsection{Human EEG-aligned ResNet also being more brain-like}

Although we trained ReAlnets based on CORnet and confirmed that they are more human brain-like, we also wondered whether our multi-layer encoding-based alignment framework could be extended to other models. Therefore, we chose ResNet18, a relatively larger model, and aligned it with the EEG representations of ten subjects from THINGS EEG2 dataset using the same framework as above. We refer to the aligned model based on pretrained ResNet18 as ReAlnet-R. Subsequently, we tested ReAlnet-Rs for their similarity to human EEG, fMRI, and behavior, comparing the results with those of the purely image-trained ResNet18.

\begin{figure}[ht]
\begin{center}
\centerline{\includegraphics[width=\columnwidth]{Figure6.jpg}}
\caption{Similar improvements in \textcolor{blue}{ReAlnet-Rs}. (A) Representational similarity time courses between human EEG and models (ReAlnet-Rs and ResNet) for layer 1, 5, 9, 13, and 17 respectively. Black square dots at the bottom indicate the timepoints where ReAlnet-Rs vs. ResNet were significantly different ($p<.05$). Lines and shading reflect mean±SEM. (B) Time courses of the maximum representational similarity between human EEG and different models (ReAlnet-Rs and ResNet), computed by taking the highest similarity across all model layers at each timepoint. Black square dots at the bottom indicate timepoints where ReAlnet-Rs significantly outperformed ResNet ($p<.05$). Lines and shading reflect mean±SEM. (C) ReAlnet-R individual variability matrices of layer 1, 5, 9, 13, and 17 and individual variability along layers. Each circle dot indicates a pair of two personalized ReAlnets. (D) Representational similarity between models and human fMRI of five different brain regions when Subject 2 in Shen fMRI test dataset viewed natural, artificial shape, and alphabetical letter images. Black asterisks indicate significantly higher similarity of ReAlnet-Rs than that of ResNet ($p<.05$). Grey asterisks indicate significantly lower similarity of ReAlnet-Rs than that of ResNet ($p<.05$). Each circle dot indicates an individual ReAlnet-R. Error bar reflects ±SEM. (E) Similarity between models and human behavior based on the Brain-Score platform. Each circle dot indicates an individual ReAlnet-R. Error bar reflects ±SEM.}
\label{Figure6}
\end{center}
\end{figure}

Firstly, ReAlnet-Rs show significantly higher similarity to human EEG neural dynamics compared to ResNet for nearly all visual layers (Layer 1: 70-160ms, Layer 5: 60-200ms, Layer 9: 60-200ms, Layer 13: 60-180ms, Layer 17: 70-160ms, \Cref{Figure6}A; Maximum similarity across all layers: 120-180ms, \Cref{Figure6}B; See all layers' EEG similarity results in \Cref{FigureS8}). Secondly, personalized ReAlnet-Rs, similar to ReAlnets, exhibit individual variability increasing with the depth of the layers (\Cref{Figure6}C; See all layers' individual variability matrices in \Cref{FigureS9}). Thirdly, ReAlnet-Rs also show higher similarity to human fMRI representations across multiple visual ROIs and different image categories (\Cref{Figure6}D; See fMRI similarity results on all three subjects in Shen fMRI dataset in \Cref{FigureS10} and detailed statistical results including t-value, p-value, and Cohen’s d are listed in \Cref{TableS4}.). Fourthly, for human behavioral similarity, although it is not significant on the average score ($t=1.6529, p=.1328, d=.5227$), seven out of ten ReAlnet-Rs show higher similarity than original ResNet. And ReAlnet-Rs show significantly higher behavioral similarity on Geirhos2021 task. These results collectively indicate that our alignment framework can be successfully extended to other visual models, such as ResNet, with ReAlnet-Rs still demonstrating improved similarity to human neural and behavioral representations.

\subsection{ReAlnets trained across subjects exhibit higher similarity to EEG and fMRI but not behavior}

While the core focus of our study is on individualized ReAlnets that capture subject-specific neural representational patterns, we also evaluated the generalization capability of our alignment framework beyond individual tuning. To this end, we trained an additional model, ReAlnet-AcrossSub, using EEG data pooled across all ten participants in the THINGS EEG2 training set. Concretely, we pooled the EEG data across subjects by treating all trials as coming from a single "super-subject", without averaging across them, and the model was trained without subject identifiers. Thus, instead of 10 individualized ReAlnets, we have a single ReAlnet-AcrossSub. Analogous to the single CORnet, we then conducted RSA by comparing the across-subject model against individual subjects' EEG and fMRI RDMs, as in our main analyses.

ReAlnet-AcrossSub shows significantly higher similarity to human EEG and fMRI representations when human participants viewed natural images compared to CORnet (\Cref{FigureS11}A-C). These results demonstrate that this across-subject model learned unified neural representations across individuals and generalized across neural modalities. However, ReAlnet-AcrossSub did not show model-fMRI similarity improvement on artificial shape or alphabetical letter images. And they did not improve behavioral similarity, performing slightly worse than CORnet (\Cref{FigureS11}D). This dissociation implies that individual-specific neural tuning may be important for improving more brain-like representations of shape or letter visual inputs and capturing human behavioral patterns. We additionally compared the across-subject ReAlnet with the individualized ReAlnets trained separately for each subject (\Cref{FigureS12}). Notably, the relative performance of model-EEG and model-fMRI similarity between them varied slightly across conditions, with individualized ReAlnets performing better in some cases (model-EEG similarity, model-fMRI similarity of shape and letter images, and model-behavior similarity) and the across-subject ReAlnet performing better in others (model-fMRI similarity of natural images).

\newpage

\section{Discussion}

Building upon previous research utilizing neural data for aligning object recognition models, we propose a novel framework for human neural representational alignment, along with the corresponding human brain-like model, ReAlnet. Unlike previous studies that focused on using animal neural signals to optimize models or were unable to use global neural activity for comprehensive model optimization \cite{Dapello2023, Federer2020, Li2019, Pirlot2022, Safarani2021}, our approach leverages human EEG activity to simultaneously optimize multiple layers of the model, enabling it to learn the human brain’s internal representational patterns for object visual processing. Notably, unlike prior research relying on behavioral or single modality neural recording data for model evaluation \cite{Dapello2023, Federer2020, Fu2023, Li2019, Pirlot2022, Safarani2021}, we employed different modalities of human neuroimaging data and also human behaviors for model evaluation to ensure that ReAlnets learn broader, cross-modal brain representational patterns. Additionally, we have extended our alignment framework to another convolutional neural network model to obtain ReAlnet-Rs and observed a similar enhancement in the similarity to human brain representations.

Recent advances in brain-inspired AI have explored various approaches to enhancing biological alignment, including self-supervised learning, recurrent architectures, and direct optimization of similarity to neural data. Unlike prior studies that primarily focus on invasive recordings from non-human primates, ReAlnets directly incorporate non-invasive human EEG data, offering a more accessible alternative for modeling human visual representations. While previous work has attempted to improve model-to-brain alignment at a single layer or specific brain region, our multi-layer encoding-based alignment framework enables a more comprehensive alignment across the visual processing hierarchy. This framework is particularly useful in contexts where biological plausibility is a key objective, such as cognitive neuroscience and brain-inspired AI. Recent studies in cognitive neuroscience have begun to apply CNNs to obtain visual features from complex stimuli \cite{McMahon2023,Bao2020,Jagadeesh2022}. Building on this, more brain-like models can now be used to extract representations that are closer to human neural patterns. This approach offers a new way to probe how neural representations envolve across different modalities (e.g., EEG vs. fMRI), stimuli, tasks, and individuals, thereby providing insights into the mechanisms underlying visual perception.

Regarding ReAlnets themselves, they effectively learn not just the patterns of EEG data, but appear to capture something even broader about the brain’s internal processing patterns of visual information. The fact that ReAlnets show higher similarity than the original CORnet not only to within-modality EEG but also to cross-modality fMRI and behavior suggests that the learned representations in ReAlnets capture shared neural patterns that are consistent across different neuroimaging modalities. One possibility is that these shared patterns may reflect the encoding of fundamental visual features, from lower- to higher-level, that are robustly represented in the brain irrespective of the neuroimaging method. In essence, the EEG data could be capturing a core, image-specific representation that generalizes across modalities, providing crucial cues for perceptual inference. And such cross-modal generalization highlights the potential of ReAlnet alignment framework as a flexible framework for extracting the key neural representations beyond single modalities. Our control experiments also highlights that only the model trained on image-specific EEG signals and including both MSE and contrastive learning losses shows this generalization performance. If we remove the contrastive learning loss, we can still see significant model-fMRI similarity improvement but not as great as ReAlnets. If we remove the MSE loss or we employed shuffled (unpaired or time-scrambled) EEG signals to train models, the model-fMRI similarity became even worse than that of the original CORnet. Future work could further explore the specific factors that drive this generalization, such as shared representational structures in EEG and fMRI neural features. And extending this alignment framework to directly incorporate fMRI data or combining fMRI and EEG for joint training could further enhance the model's ability to capture cross-modal brain representations. Additionally, the ability of ReAlnets to generalize from EEG to fMRI suggests a degree of representational consistency across neural modalities. This observation raises the possibility that different neuroimaging techniques, despite capturing distinct aspects of neural activity (e.g., temporal dynamics in EEG vs. spatial patterns in fMRI), may share a common representational basis \cite{Cichy2020,LeeMasson2023,Hu2025}. Investigating this consistency further could provide insights into the shared neural mechanisms underlying human visual processing and inform the design of more versatile brain-aligned models.

Recent findings have highlighted the critical role of the training image diet in determining the model-to-brain fit \cite{Conwell2024}. In addition to the generalization to other neuroimaging modalities, our ReAlnets trained on THINGS EEG2 training dataset also demonstrated robust generalization to held-out THINGS categories. This generalization suggests that the networks can capture shared representational structures that are meaningful both within and beyond the training distribution. Such robustness indicates that training on semantically rich and ecologically valid images and the corresponding brain signals, such as THINGS EEG2, may promote the emergence of brain-like visual representations that generalize across diverse visual domains.  On the one hand, our findings demonstrate that our approach improves model-brain similarity. On the other hand, these results may also suggest that selecting appropriate training datasets might be important for letting artificial models learn generalized human brain representations.

EEG signals provide high temporal resolution data that capture rapid neural dynamics underlying visual processing. By leveraging EEG data, our alignment framework enables the model to align with temporal features that likely correspond to distinct stages of visual processing, such as early sensory features and later semantic attributes. Also, compared to fMRI or electrophysiology, EEG is significantly more cost-effective, making it more practical for widespread use. Recent studies in cognitive and computational neuroscience using the THINGS EEG2 dataset have traced evidence of human visual feature processing, including object categories, size, depth, image entropy \cite{Lu2023d, Muukkonen2023}, and even reconstructing visual information through EEG signals \cite{Li2024, Du2023, Song2023} and realizing inter-subject EEG conversions \cite{Lu2023c}. Our internal representational analysis suggests that ReAlnets contain enhanced representations of object dimensions, such as food-related, artificial, electronic, etc. These findings suggest that human brain EEG signals may capture aspects of visual information not present in traditional convolutional neural networks. Similar to this perspective, the fact that different generation loss weights do not significantly impact the model category classification performance but do enhance its similarity to human brains suggests that nodes in the model, which originally did not encode category-specific information, may have been optimized \cite{Federer2020}. Also, to address the potential confound that changes in visual “diet” — such as an increased prevalence of food-related items in THINGS compared to ImageNet — may underlie the observed representational shifts, we conducted a control experiment in which all food-related image-EEG pairs were removed during alignment training. Even without exposure to food-related stimuli or neural signals, ReAlnets still exhibited enhanced food-related representations, though to a lesser extent. This suggests that neural alignment, not stimulus statistics alone, is the primary driver of representational changes. While this experiment specifically targeted the food dimension, the findings may reflect a general trend: ReAlnets captures abstract, brain-like representational structure present in neural data, beyond what can be accounted for by training image distributions. However, it does warrant further exploration to ascertain what specific information has been learned from the alignment with human brains. More analyses of the neural network’s internal representations may be needed to delve into this. Also, from a reverse-engineering perspective, attempting to understand the brain-like optimization process of the model could further aid in unraveling the mechanisms by which our brains process visual information \cite{Ayzenberg2023, Cichy2019, Doerig2022, Kanwisher2023, Lu2023}.

Interestingly, when we applied this alignment framework to ResNet18, the resulting ReAlnet-Rs still demonstrate more human brain-like representations, akin to those exhibited by ReAlnets. This framework's generalizability may suggest that our alignment framework is potentially applicable to aligning other AI models with human EEG signals. Therefore, one potential direction for future research is to examine whether this alignment framework could generalize to other model architectures or neuroimaging modalities. First, while the current study only utilizes EEG signals for alignment, the observed generalization to fMRI and behavior suggests potential cross-modal consistency in the learned representations. However, we acknowledge that our findings do not directly demonstrate generalization from fMRI to EEG. Future work incorporating direct fMRI-based alignment or joint EEG–fMRI training would be necessary to establish the full bidirectionality of cross-modal generalization. Future studies could test whether this alignment framework could be extended to other neural modalities, such as fMRI and MEG (dimensionality reduction might be necessary for extensive neural data features), suggesting potential applicability of this framework to variants aligned with fMRI or MEG data. Second, another ambition is to adapt this framework to a wider range of models and tasks in the future, including language and auditory processing and self-supervised or unsupervised models, which may be adapted for other modalities and learning paradigms, including auditory or language inputs and unsupervised learning settings. Third, one direction for refinement includes integrating loss functions designed to emphasize task-relevant neural features. Also, although our framework successfully demonstrates generalization to fMRI data, further work is needed to systematically analyze the factors contributing to this transfer. For instance, identifying representational dimensions that are consistent across EEG and fMRI, and relating them to specific brain regions, could provide deeper insights into the mechanisms of cross-modal generalization.

While this study demonstrates encouraging progress, several limitations warrant consideration. First, the relatively small sample size of EEG data and the high level of signal noise may limit the precision of model-to-brain alignment. Second, the lack of shared category labels across datasets—such as the absence of ImageNet labels for THINGS stimuli—complicates consistent model evaluation. These issues may constrain the model's performance both in brain alignment and object recognition. In addition, although ReAlnets were fine-tuned on EEG signals from individual subjects, they still inherit architectural and representational biases from the pretrained image recognition model, which may limit alignment fidelity. Beyond these data-related factors, certain limitations may also stem from the alignment methodology itself. For instance, our framework employs relatively simple alignment objectives—such as MSE and contrastive losses—that may not fully capture the complex, nonlinear, and distributed nature of brain representations. Furthermore, the shallow encoding modules assume a direct mapping between neural activity and model features, potentially overlooking intermediate transformations or multistage processing. These design choices could limit the model’s capacity to learn more abstract or hierarchically structured neural patterns. Thus, the modest effect sizes observed may reflect not only data constraints, but also the current methodological limits of representational alignment strategies.

Despite these limitations, the framework consistently improves alignment with human EEG, fMRI, and behavior. To build on this work, future efforts could assess whether training the model jointly on EEG and fMRI improves cross-modal generalization, and whether the framework remains robust when using smaller or noisier EEG datasets. Moreover, evaluating how alternative training losses or network architectures impact representational similarity may reveal which components are critical for capturing brain-like representations. Finally, although the THINGS EEG2 dataset is ecologically valid, further work is needed to test whether these findings generalize to datasets with different semantic or contextual structures.

Overall, this study underscores the importance of investigating the limitations and potentials of human EEG signals in shaping brain-like AI.  We employ a novel alignment framework using human EEG data to achieve more human brain-like vision models - ReAlnets. Demonstrating significant advances in bio-inspired AI, ReAlnets not only align closely with human EEG and fMRI but also exhibit hierarchical individual variability and increased similarity to human behavior, mirroring human visual processing. We hope that our alignment framework stands as a testament to the potential synergy between computational neuroscience and machine learning and enables the enhancement of diverse AI models to be more human brain-like, opening up exciting possibilities for future research in brain-like AI systems.

\section{Methods}

Here we describe the human neural data (EEG data for the alignment, and both EEG and fMRI data for testing the similarity between models and human brains) we used in this study, the alignment pipeline (including the structure, the loss functions, and training and test methods) for aligning the model representations with human neural representations, and the evaluation methods for measuring representational similarity between models and human brains and human behaviors.

\subsection{Human EEG data for representational alignment}

Human EEG data were obtained from an EEG open dataset, THINGS EEG2 \cite{Gifford2022}, including EEG data from 10 healthy human subjects in a rapid serial visual presentation (RSVP) paradigm. Stimuli were images sized 500 $\times$ 500 pixels from THINGS dataset \cite{Hebart2019}, which consists of images of objects on a natural background from 1854 different object concepts. Before imputing the images to the model, we reshaped image sizes to 224 $\times$ 224 pixels and normalized the pixel values of images to ImageNet statistics. Subjects viewed one image per trial (100ms). Each participant completed 66160 training set trials (1654 object concepts $\times$ 10 images per concept $\times$ 4 trials per image) and 16000 test set trials (200 object concepts $\times$ 1 image per concept $\times$ 80 trials).

EEG data were collected using a 64-channel EASYCAP and a BrainVision actiCHamp amplifier. We use already pre-processed data from 17 channels (O1, Oz, O2, PO7, PO3, POz, PO4, PO8, P7, P5, P3, P1, Pz, P2) overlying occipital and parietal cortex. We re-epoched EEG data ranging from stimulus onset to 200ms after onset with a sample frequency of 100Hz. Thus, the shape of our EEG data matrix for each trial is 17 channels $\times$ 20 time points. and we reshaped the EEG data as a vector including 340 values for each trial. Before the model training and test, we averaged all the repeated trials (4 trials per image in the training set and 80 trials per image in the test set) to obtain more stable EEG signals.

It is worth noting that the training and test sets do not overlap in terms of object categories (concepts), which means that the performance of ReAlnets trained on the training set, when evaluated on the test set, can effectively reveal the model’s generalization capability across different object categories.

\subsection{Human fMRI data for cross-modality testing}

To demonstrate that our approach of aligning with human EEG not only enhances the model's similarity to human EEG but indicates that ReAlnets have effectively learned the human brain's representational patterns more broadly, we also performed cross-modal testing, testing ReAlnets on data from a different modality (fMRI), from a different set of subjects, viewing a different set of images. The fMRI data originate from \cite{Shen2019}. This \textit{Shen fMRI dataset} recorded human brain fMRI signals from three subjects while they focused on the center of the screen viewing images. We selected the test set from \textit{Shen fMRI dataset}, which comprises fMRI signals of each subject viewing 50 natural images of different categories from ImageNet, 40 artificial shape images, and 10 alphabetical letter images with each image being viewed 24, 20, and 12 times respectively. We averaged the fMRI signals across the repeated trials to obtain more stable brain activity for each image observation and extracted signals from five regions-of-interest (ROIs) for subsequent comparison of model and human fMRI similarity: V1, V2, V3, V4, and the lateral occipital complex (LOC).

\subsection{Image-to-brain encoding-based alignment pipeline}

\textbf{Basic architecture of ReAlnets and ReAlnet-Rs}: We have chosen the state-of-the-art CORnet-S model \cite{Kubilius2018, Kubilius2019} as the foundational architecture for ReAlnets, incorporating recurrent connections akin to those in the biological visual system and proven to more closely emulate the brain’s visual processing. Both CORnet and ReAlnets consist of four visual layers (V1, V2, V4, and IT) and a category decoder layer. Layer V1 performs a 7 $\times$ 7 convolution with a stride of 2, followed by a 3 $\times$ 3 max pooling with a stride of 2, and another 3 $\times$ 3 convolution. Layer V2, V4, and IT each perform two 1 $\times$ 1 convolutions, a bottleneck-style 3 $\times$ 3 convolution with a stride of 2, and a 1 $\times$ 1 convolution. Apart from the initial Layer V1, the other three visual layers include recurrent connections, allowing outputs of a certain layer to be passed through the same layer several times (twice in Layer V2 and IT, and four times in Layer V4). We have also chosen another widely used model in image recognition, ResNet18, as the foundational architecture to obtain human EEG-aligned ReAlnet-Rs, which consists of 18 layers (the last layer is the final decoder to output the predicted category label). 

\noindent\textbf{EEG generation module:} For ReAlnets, in addition to the original recurrent convolutional neural network structure, we have added an EEG generation module designed to construct an image-to-brain encoding model for generating realistic human EEG signals. Each visual layer is connected to a nonlinear \textit{N} $\times$ 128 layer-encoder (Enc-V1, Enc-V2, Enc-V4, and Enc-IT correspond to Layer V1, V2, V4, and IT) that processes through a fully connected network with a ReLU activation. These four layer-encoders are then directly concatenated to form an \textit{N} $\times$ 512 Multi-Layer Visual Encoder, which is subsequently connected to an \textit{N} $\times$ 340 EEG encoder through a linear layer to generate the predicted EEG signals. Here \textit{N} is the batch size. For eAlnet-Rs which is highly similar to ReAlnets, we extracted the features from Layer 5, Layer 9, Layer 13, and Layer 17 to connect to four nonlinear \textit{N} $\times$ 128 layer-encoders through fully connected networks with ReLU activations, and these four layer-encoders are then directly concatenated to form an \textit{N} $\times$ 512 Multi-Layer Visual Encoder, which is subsequently connected to an \textit{N} $\times$ 340 EEG encoder through a linear layer to generate the predicted EEG signals.

Therefore, we aim for the model to not only perform the object classification task but also to generate human EEG signals which can be highly similar to the real EEG signals when a person views the certain image through the EEG generation module with a series of encoders. During this process of generating brain activity, ReAlnet(-R)s' visual layers are poised to effectively extract features more aligned with neural representations.

\noindent\textbf{Alignment Loss:} Accordingly, the training loss $\mathcal{L}^A$ of our alignment framework consists of two primary losses, a classification loss and a generation loss with a parameter $\beta$ that determines the relative weighting:

\begin{equation}
\mathcal{L}^A=\mathcal{L}^C+ \beta \cdot \mathcal{L}^G
\end{equation}

$\mathcal{L}^C$ represents the standard categorical cross entropy loss for model predictions on ImageNet labels:

\begin{equation}
    \mathcal{L}^C=-\sum_{i=1}^N y_i log(p_i)
\end{equation}

Here, $y_i$ represents the $i$-th image, and $p_i$ represents the probability that model predicts the $i$-th image belongs to class $i$ out of 1000 categories. However, the correct ImageNet category labels for images in THINGS dataset are not available. Therefore, we adopt the same strategy as in \cite{Dapello2023}, using the labels obtained from the ImageNet pre-trained CORnet without neural alignment as the true labels to stabilize the classification performance of ReAlnets.

$\mathcal{L}^G$ is the generation loss, which includes a mean squared error (MSE) loss $\mathcal{L}^{MSE}$ and a contrastive loss $\mathcal{L}^{Cont}$ between the generated and real EEG signals. To compute the contrastive loss, we originally aimed to use Spearman correlation to measure the similarity between predicted and target signals. However, as Spearman correlation involves a ranking operation that is non-differentiable, we replaced it with Pearson correlation, which is a differentiable similarity metric. Specifically, the dissimilarity index was calculated as 1 minus Pearson correlation. This substitution allowed us to compute gradients effectively during backpropagation, ensuring the compatibility of the contrastive loss with gradient-based optimization. The contrastive loss aims to bring the generated signals from the same image (positive pairs) closer to the corresponding real human EEG signals and make the generated signals from different images (negative pairs) more distinct. $\mathcal{L}^G$ is calculated as followed:

\begin{equation}
    \mathcal{L}^G = \mathcal{L}^{MSE} + \mathcal{L}^{Cont}
\end{equation}
\begin{equation}
    \mathcal{L}^{MSE} = \frac{1}{N}\sum_{i=1}^N(S_i - \hat{S_i})^2
\end{equation}
\begin{equation}
    \begin{split}
    \mathcal{L}^{Cont} = 1+\frac{1}{N}\sum_{i=1}^N[1-r(S_i, \hat{S_i})] \\ - \frac{1}{N(N-1)}\sum_{i=1}^N\sum_{j=1, j\ne i}^N[1- r(S_i, \hat{S_j})]
    \end{split}
\end{equation}

Here, $S_i$ and $\hat{S}_i$ represent the generated and real EEG signals corresponding to the $i$-th image.

\noindent\textbf{Training procedures:} Unlike CORnet and ResNet18 which trained on purely image-based ImageNet dataset, ReAlnets and ReAlnet-Rs additionally trained on individual EEG data. According to ten subjects in THINGS EEG2 dataset, we obtained ten personalized ReAlnets. Each network was trained to minimize the alignment loss including both classification and generation losses with a static loss weight $\beta$ of 100 and a static training rate of 0.00002 for 30 epochs using the Adam optimizer. We used a batch size of 16, meaning the contrastive loss computed dissimilarities of 256 pairs for each gradient step.

Additionally, for ReAlnets, we applied other three different $\beta$ weights ($\beta$ = 1, 10, or 1000) separately to train the model to further explore the impact of this $\beta$ value on the performance of ReAlnets. We observed that with an increase in $\beta$, ReAlnets show greater similarity to human EEG and fMRI and more pronounced individual variability within models. However, only ReAlnets with $\beta$ = 100 show significantly higher similarity to human behaviors. Thus, we suggest that $\beta$ = 100 could be the best parameter to conduct the human EEG alignment. \Cref{FigureS13,FigureS14,FigureS15,FigureS16} show the performance and similarity results of ReAlnets with different $\beta$ values.

We tested the classification accuracy of ReAlnets on ImageNet at different $\beta$ values (\Cref{FigureS1}A). Importantly, to ascertain that the observed decrease in accuracy was not due to the additional generation task compromising classification performance, but rather the absence of correct ImageNet labels for images in THINGS EEG2 dataset, we trained a ReAlnet with $\beta$ = 0. This ReAlnet excluded the EEG signal generation module but underwent fine-tuning with images from THINGS EEG2 dataset. The results indicated that the ReAlnet with $\beta$ = 0 also experienced a similar level of decline.

\noindent\textbf{Control experiments:} To systematically evaluate how different losses and how shuffled EEG data influence our results we conducted four control experiments. The corresponding four control conditions are defined as follows: (1) \textit{W/o ContLoss} control: We removed the constrastive loss component; (2) \textit{W/o MSELoss} control: We removed the MSE loss component; (3) Unpaired EEG control: We randomly reassigned EEG signals to images, breaking the image-EEG pairing while preserving the temporal structure of the EEG data; (4) Scrambled EEG control: We temporally permuted the EEG signal within each pair, disrupting the temporal structure of the EEG signal while preserving its overall distribution.

\subsection{Model-human similarity measurement}

\noindent\textbf{Neural similarity via Representational similarity analysis (RSA):} RSA is used for representational comparisons between models and human brains \cite{Kriegeskorte2008a} based on first computing representational dissimilarity matrices (RDMs) for models and human neural signals, and then calculating Spearman correlation coefficients between RDMs from two systems.

To evaluate the similarity between models and human EEG, we calculated EEG RDMs using classification-based decoding accuracy as the dissimilarity index. While fMRI RDMs are typically calculated using 1 minus correlation coefficient (below) \cite{Kriegeskorte2008a,Nili2014,Cichy2014,Cichy2016}, decoding accuracy is more commonly used for EEG RDMs \cite{Grootswagers2017,Xie2020}. Since EEG has a low SNR and includes rapid transient artifacts, Pearson correlations computed over very short time windows yield unstable dissimilarity estimates \cite{Kappenman2010,Luck2014} and may thus fail to reliably detect differences between images. In contrast, decoding accuracy - by training classifiers to focus on task-relevant features - better mitigates noise and highlights representational differences. For each image, we extracted the corresponding EEG response matrix (80 trials $\times$ 17 channels $\times$ 20 time-points). For each subject and each time-point, we referred 80 trials per image as training samples and 17 channels as features, resulting in a total of 160 samples for every pair of two images \textit{i} and \textit{j}. A linear SVM classifier was employed to classify the EEG responses between two images using a 5-fold cross-validation approach. In each fold the classifier was trained on 4 folds of the data and tested on the remaining fold, ensuring that the decoding accuracy reflected generalizable patterns. The average decoding accuracy across the 5 folds was computed and used as the dissimilarity measure between images \textit{i} and \textit{j}. And this process was repeated for all possible pairs of the 200 images, resulting in a 200 × 200 RDM for each subject and each time point. For model RDMs, we input 200 images into each model and obtained latent features from each visual layer. Then, we constructed each layer’s RDM by calculating the dissimilarity using 1 minus Pearson correlation coefficient between flattened vectors of latent features corresponding to any two images. To compare the representations, we calculated the Spearman correlation coefficient as the similarity index between layer-by-layer model RDMs and timepoint-by-timepoint neural EEG RDMs.

To evaluate the similarity between models and human fMRI, we used the Shen fMRI dataset, which includes fMRI activation patterns for different image categories (natural images, artificial shape images, and alphabetical letter images). We calculated separate RDMs for each category, and the RDM dimensions correspond to the number of images in each category: 50 $\times$ 50 for natural images, 40 $\times$ 40 for artificial shape images, and 10 $\times$ 10 for alphabetical letter images. We used the GLM activation values (beta weights) provided in the open dataset for each image and each voxel, and calculated RDMs based on the common correlation-based method \cite{Kriegeskorte2008a,Nili2014,Cichy2014,Cichy2016}, as follows. For each subject and ROI, we defined the voxel-wise activation pattern for each image as the vector of activation values (e.g., if a given ROI had 250 voxels, this would be a 1$\times$250 vector for each image). Then, for every pair of images, we computed the dissimilarity index as 1 minus Pearson correlation coefficient between the two voxel-wise activation vectors corresponding to the two images. This was repeated for all image pairs, resulting in a symmetrical dissimilarity matrix (the RDM) for each ROI and each subject. For model RDMs, similar to the EEG comparisons above, we obtained the RDM for each layer from each model. Then, we calculated the Spearman correlation coefficient as the similarity index between layer-by-layer model RDMs and neural fMRI RDMs for different ROIs, assigning the final similarity for a certain brain region as the highest similarity result across model layers due to the lack of a clear correspondence between different model layers and brain regions. All RSA analyses were implemented based on NeuroRA toolbox \cite{Lu2020}.

\noindent\textbf{Behavioral similarity via Brain-Score:} 

Brain-Score is a framework that evaluates how similar artificial neural networks (ANNs) are to the primate visual system \cite{Schrimpf2020}. To measure the behavioral similarity between ReAlnets and humans (and monkeys) in visual recognition tasks, we used two behavioral benchmarks from the Brain-Score framework:(\url{https://github.com/brain-score/vision}). "Rajalingham2018public-i2n" \cite{Rajalingham2018} task uses grayscale images where objects are manipulated by varying position, size, viewing angle, and background, while "Geirhos2021-error\_consistency" \cite{Geirhos2021} task employs out-of-distribution colorful natural images. Both tasks calculate behavioral similarity between the model and human (and primates) observers using the error consistency method, which measure whether there is above-chance overlap in the specific images that humans and models classify incorrectly. The behavioral Brain-Score is calculated by taking the average of two behavioral benchmarks. We compared the results from ReAlnets and ReAlnet-Rs to the behavioral Brain-Score of CORnet and ResNet respectively, using the same benchmarks. For more detailed information about the behavioral benchmarks used in this study, please refer to the original papers by \cite{Schrimpf2020, Rajalingham2018, Geirhos2021}. 

\noindent\textbf{Individual variability in ReAlnets:} To quantify inter-individual representational variability among the individualized ReAlnets, we computed a variability index at each model layer. Specifically, we extracted the RDM from each individualized ReAlnet (one per subject) for a given layer and then calculated all pairwise Spearman correlations between the RDMs of the ten models. The variability index was defined as the average of one minus Spearman correlation coefficient between two RDMs. A higher value indicates greater variability (i.e., less similarity) across models.

\noindent\textbf{Statistical Analysis:} 

Statistical significance was determined using two-tailed one-sample t-tests between ten ReAlnets and the single CORnet, or paired t-tests between the ten sets of ReAlnets and control models. We considered $p<.05$ as the threshold for significance. To ensure that statistical significance reflects meaningful differences, we also report effect sizes in the supplementary materials.

\subsection{Model internal representational analysis}

This section describes the methodology used for the analysis in Results - Refined object feature representations in ReAlnets section, where we examined which object feature dimensions were more strongly encoded in ReAlnets compared to CORnet. Specifically, we utilized 49 object feature dimensions from the THINGS dataset \cite{Hebart2020}. Each object concept is represented along these 49 feature dimensions. Our analysis focused on the 200 images in the test set of the THINGS EEG2 dataset, which were not part of the model's training data. We applied an RDM-based partial Spearman correlation method for the analysis. First, we computed the RDM for the IT layer of each model, which contains higher-level information, and 49 feature RDMs based on the 200 images by calculating the absolute differences in feature encoding strength between pairs of images as dissimilarity measures. Next, we computed the partial correlation between the model RDM and each feature RDM, while controlling for the other 48 feature RDMs. Finally, we calculated the square of the partial correlation coefficient to determine the variance explained by the model for each object feature dimension and got the top-3 improved feature dimensions of ReAlnets.

\section{Data availability}

The models and the analysis code can be assessed at https://github.com/ZitongLu1996/ReAlnet.

\section*{Acknowledgment}

This work was supported by grants from the National Institutes of Health (R01-EY025648) and National Science Foundation (NSF 1848939) to Julie D. Golomb. We thank the Ohio Supercomputer Center and Georgia Stuart for providing the essential computing resources and support. We thank Yuxuan Zeng for the `ReAlnet' name suggestion. We thank  Tianyu Zhang, Shuai Chen, Jiaqi Li, and some other members in Memory \& Perception Reviews Reading Group (RRG) for helpful discussions about the methods and results. We thank Yuxin Wang for constructive feedback on the manuscript.

\bibliographystyle{unsrt}
\bibliography{references}

\newpage

\setcounter{figure}{0} 
\renewcommand{\thefigure}{S\arabic{figure}}
\renewcommand{\thetable}{S\arabic{table}}

\begin{figure}[h!]
\begin{center}
\centerline{\includegraphics[width=0.9\columnwidth]{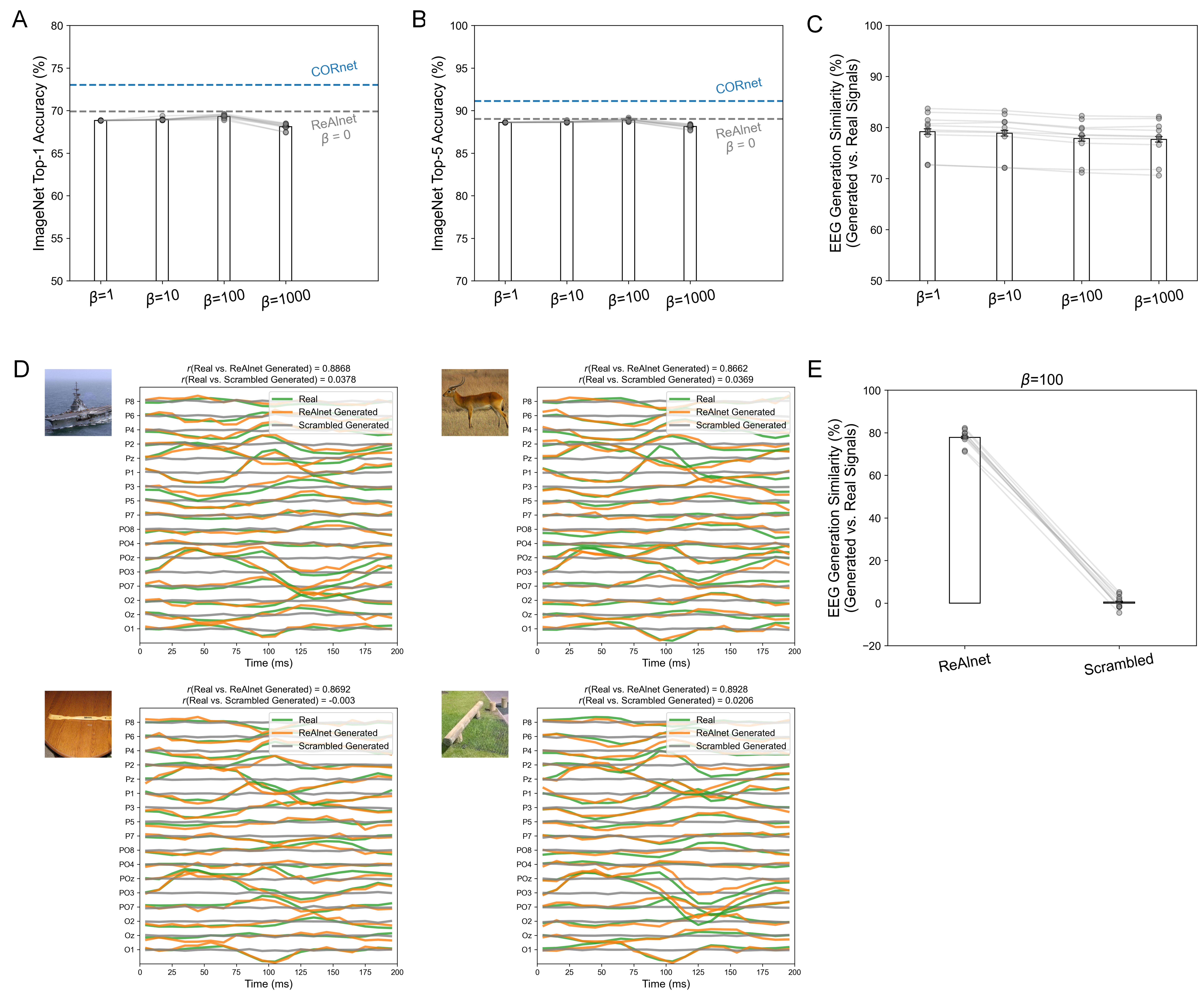}}
\caption{ReAlnets' object category classification and EEG generation performances. (A) Top-1 and (B) Top-5 ImageNet classification accuracy of different ReAlnets and CORnet. Error bar reflects ±SEM. (C) EEG generation performance (Spearman correlation between generated and real EEG signals) of different ReAlnets. Error bar reflects ±SEM. (D) Four examples of EEG generation results (from the ReAlnet and the scrambled model at $\beta$ = 100 of Sub-01). For each example, the left image indicates the image input to the models and the image viewed by the subject. The green curves represent the real EEG signals, and the orange and grey curves represent the generated EEG signals by the ReAlnet and the scrambled model corresponding to the same image. (E) EEG generation performance (Spearman correlation between generated and real EEG signals) of ReAlnets and scrambled models. Error bar reflects ±SEM. Each circle dot represent an individual ReAlnet or scrambled model.}
\label{FigureS1}
\end{center}
\end{figure}

\newpage

\begin{figure}[h!]
\begin{center}
\centerline{\includegraphics[width=0.6\columnwidth]{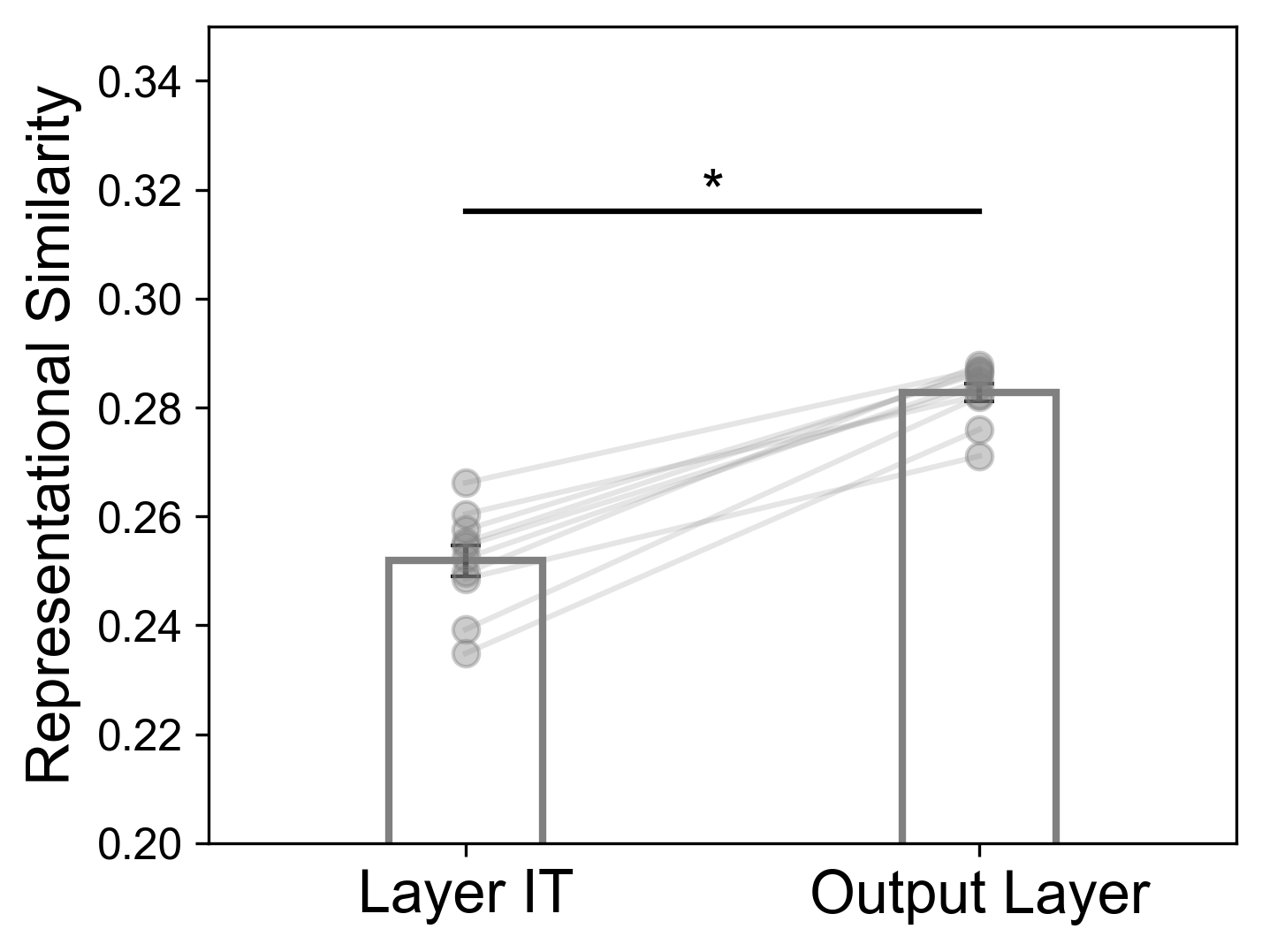}}
\caption{Representational similarity between brain LOC region (based on the Shen fMRI dataset) and the ReAlnets' Layer IT and output layer. LOC is significantly more similar to ReAlnets' output layer than Layer IT ($t=9.0272, p<.0001, d=4.0371$). Each circle dot indicates an individual model. For each model, we averaged the results across three subjects from the Shen fMRI dataset. Asterisks indicate the significance ($p<.05$). Error bar reflects ±SEM.}
\label{FigureS2}
\end{center}
\end{figure}

\newpage

\begin{figure}[h!]
\begin{center}
\centerline{\includegraphics[width=0.6\columnwidth]{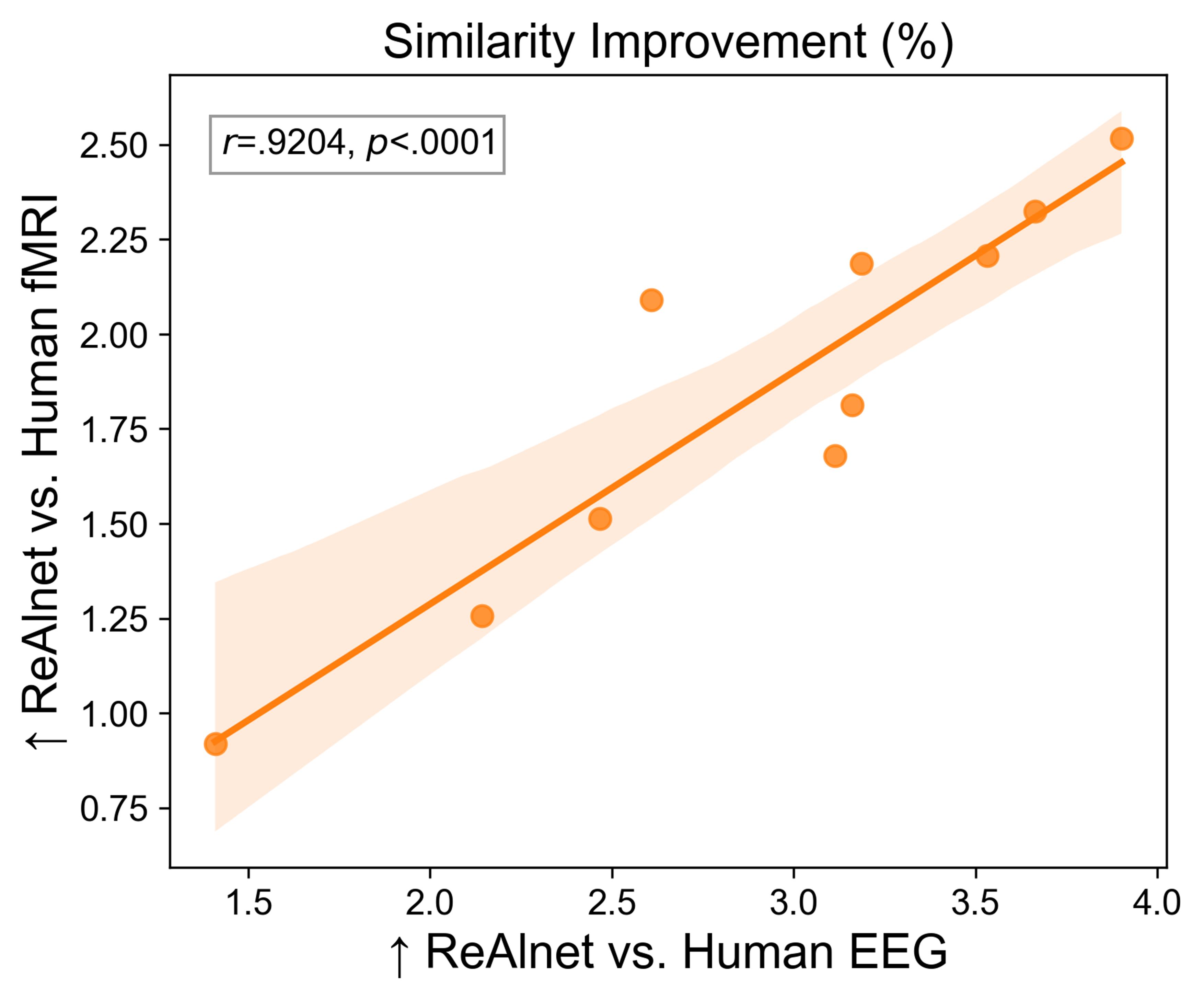}}
\caption{Scatter plot of similarity improvement between ReAlnet vs. human EEG and ReAlnet vs. human fMRI. For model-EEG similarity, we averaged the representational similarity from 50 to 200ms and across four model layers. For model-fMRI similarity, we averaged the representational similarity across five brain regions and three subjects from the Shen fMRI dataset. Each circle dot indicates an individual ReAlnet.}
\label{FigureS3}
\end{center}
\end{figure}

\newpage

\begin{figure}[h!]
\begin{center}
\centerline{\includegraphics[width=0.6\columnwidth]{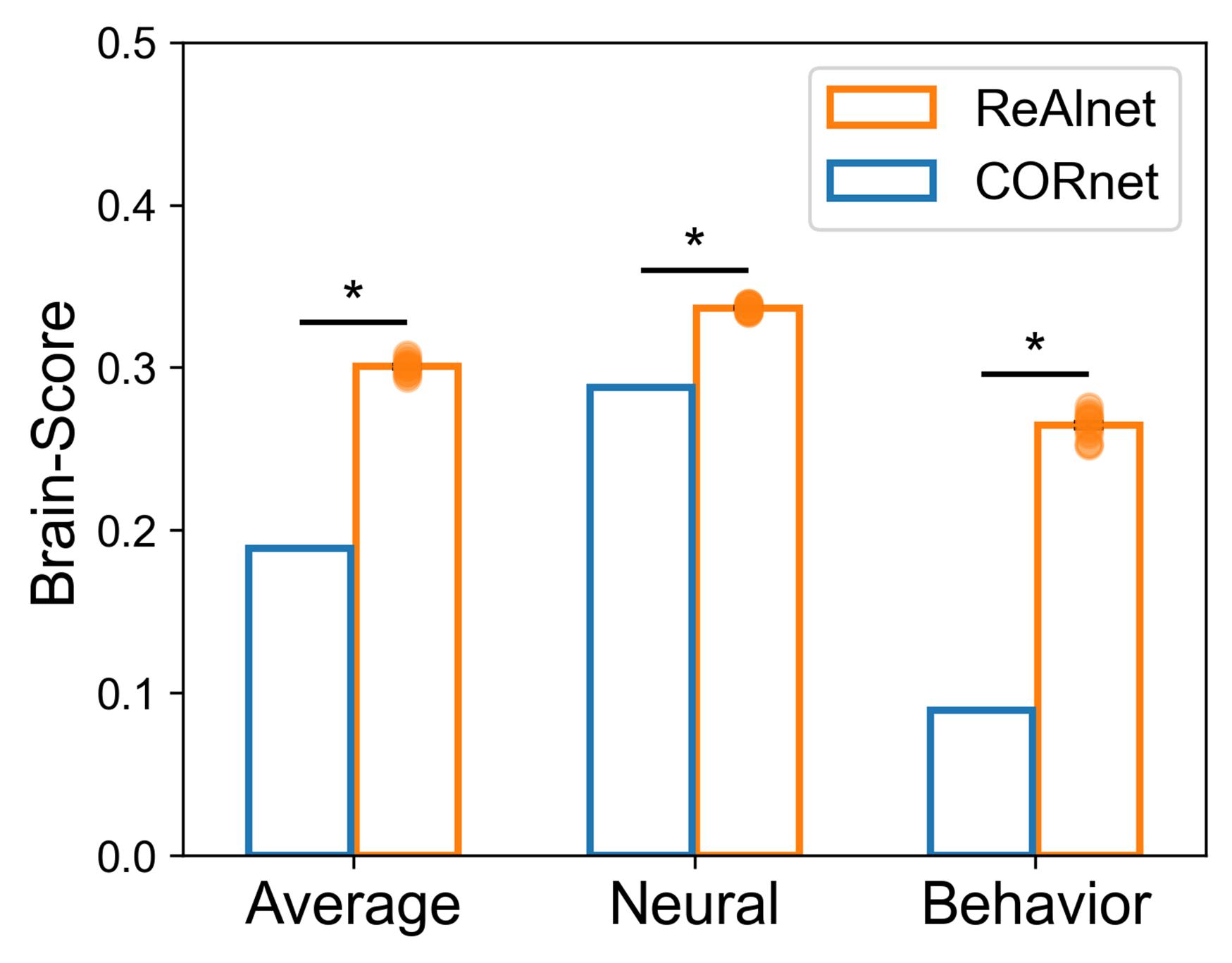}}
\caption{Brain-Score of ReAlnets and CORnet tested on updated Brain-Score platform. Asterisks indicate significantly higher similarity of ReAlnets than that of control models ($p<.05$). Error bar reflects ±SEM.}
\label{FigureS4}
\end{center}
\end{figure}

\newpage

\begin{figure}[h!]
\begin{center}
\centerline{\includegraphics[width=0.9\columnwidth]{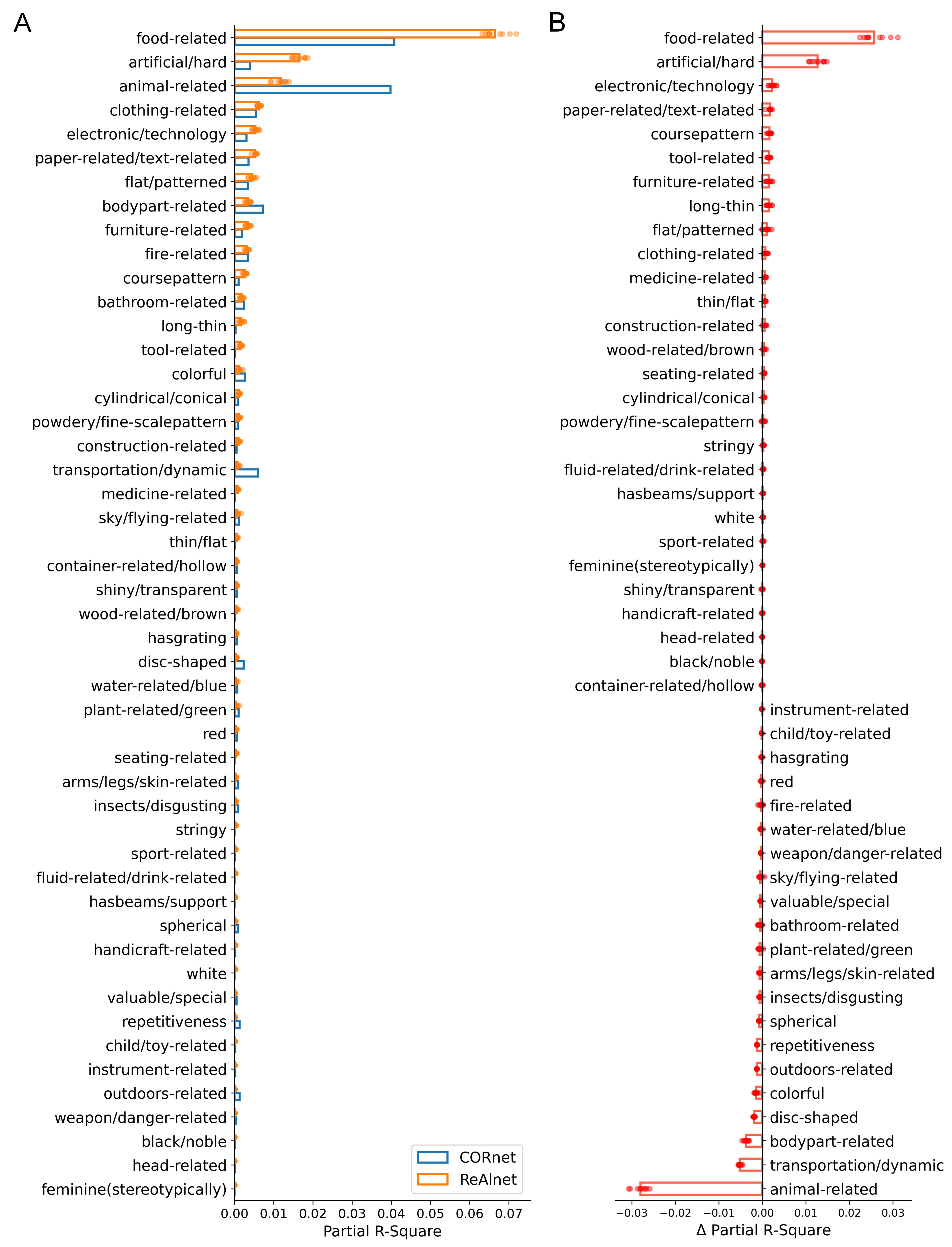}}
\caption{Internal representations in ReAlnets and CORnet. (A) Partial r-square of each object dimension in predicting the internal representations in ReAlnets and CORnet, where partial r-square is the unique variance explained by that dimension after controlling for all others. Blue bars show CORnet, and orange bars show the mean over individual ReAlnets. Dimensions are sorted by the mean partial r-square of ReAlnets. (B) The difference of partial r-square between ReAlnets and CORnet for each dimension. Dimensions are sorted by the differences to highlight which features ReAlnets gain or lose sensitivity to relative to CORnet. Each circle dot indicates an individual ReAlnet.}
\label{FigureS5}
\end{center}
\end{figure}

\newpage

\begin{figure}[h!]
\begin{center}
\centerline{\includegraphics[width=0.9\columnwidth]{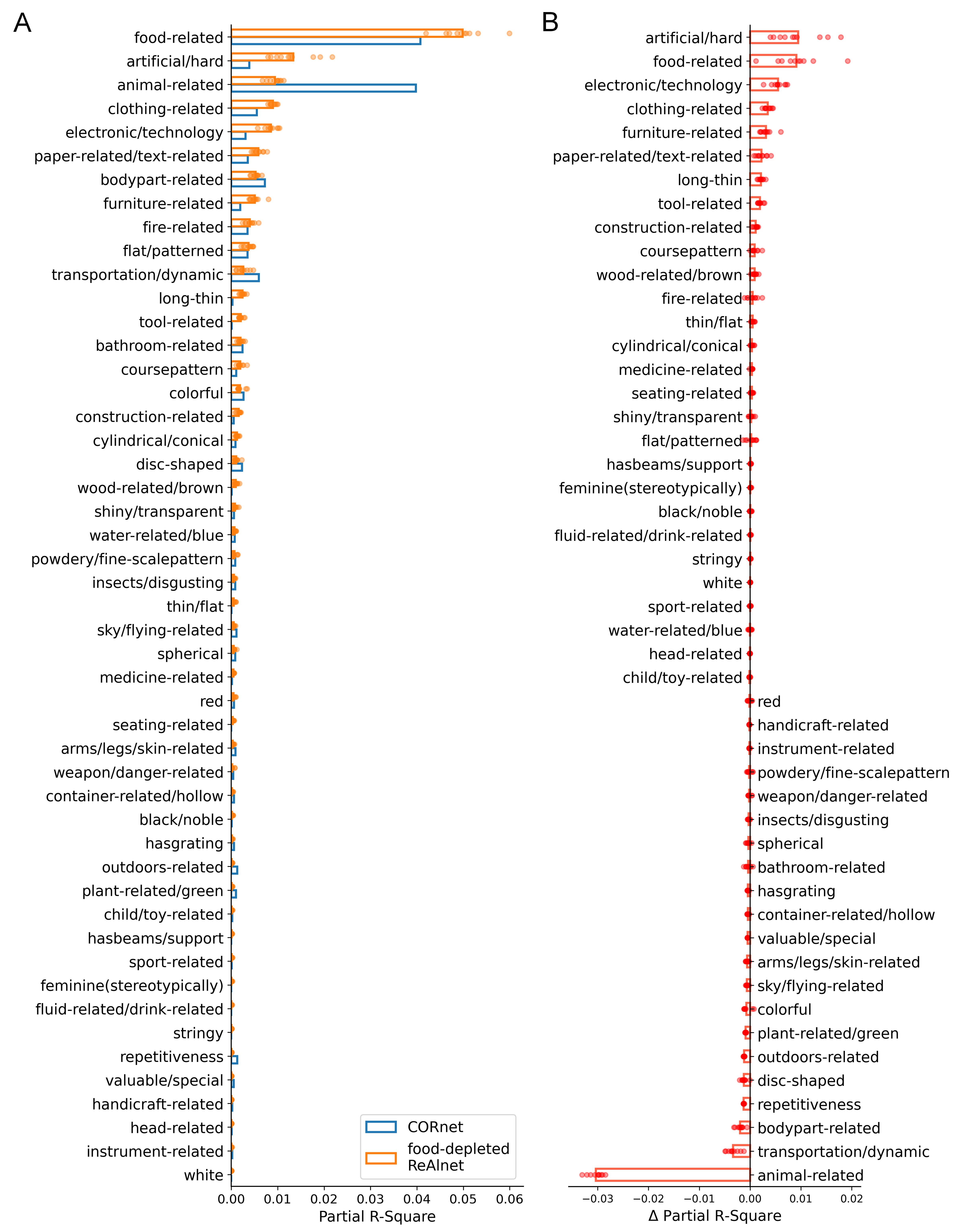}}
\caption{Internal representations in food-depleted ReAlnets and CORnet. (A) Partial r-square of each object dimension in predicting the internal representations in food-depleted ReAlnets and CORnet, where partial r-square is the unique variance explained by that dimension after controlling for all others. Blue bars show CORnet, and orange bars show the mean over individual food-depleted ReAlnets. Dimensions are sorted by the mean partial r-square of ReAlnets. (B) The difference of partial r-square between food-depleted ReAlnets and CORnet for each dimension. Dimensions are sorted by the differences to highlight which features food-depleted ReAlnets gain or lose sensitivity to relative to CORnet. Each circle dot indicates an individual food-depleted ReAlnet.}
\label{FigureS6}
\end{center}
\end{figure}

\newpage

\begin{figure}[h!]
\begin{center}
\centerline{\includegraphics[width=\columnwidth]{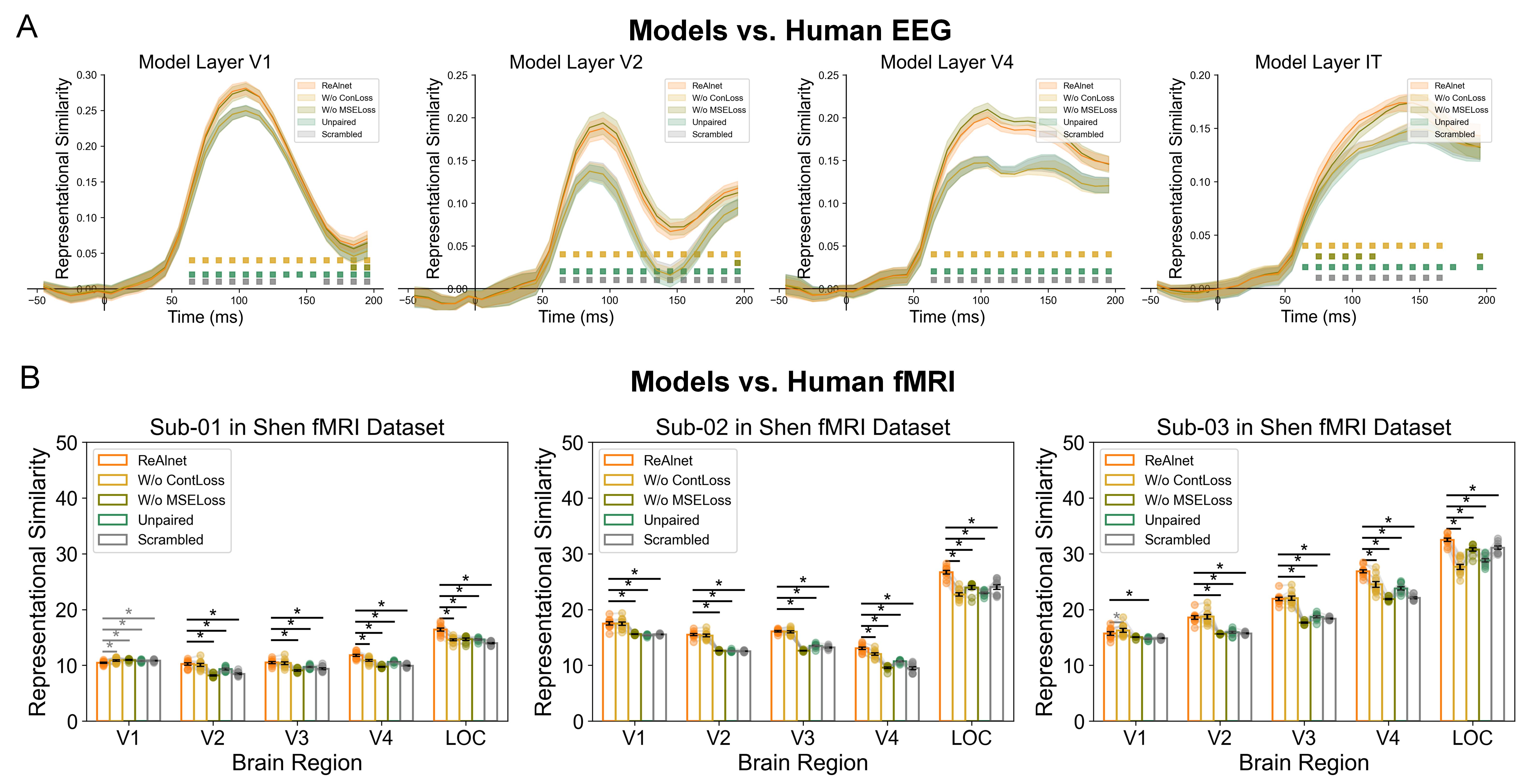}}
\caption{(A) Representational similarity time courses between human EEG and ReAlnets and control models ($\beta$ = 100) for different layers respectively. Lines and shading reflect mean±SEM. Goldenrod, olive, seagreen, and grey square dots at the bottom indicate the timepoints where ReAlnets were significantly higher than W/o ContLoss, W/o MSELoss, Unpaired, and Scrambled models respectively ($p<.05$). Lines and shading reflect mean±SEM. (B) Representational similarity between three subjects’ fMRI activity of five different brain regions and ReAlnets and control models ($\beta$ = 100) respectively. Black asterisks indicate significantly higher similarity of ReAlnets than that of control models ($p<.05$). Grey asterisks indicate significantly lower similarity of ReAlnets than that of control models ($p<.05$). Error bar reflects ±SEM.}
\label{FigureS7}
\end{center}
\end{figure}

\newpage

\begin{figure}[h!]
\begin{center}
\centerline{\includegraphics[width=\columnwidth]{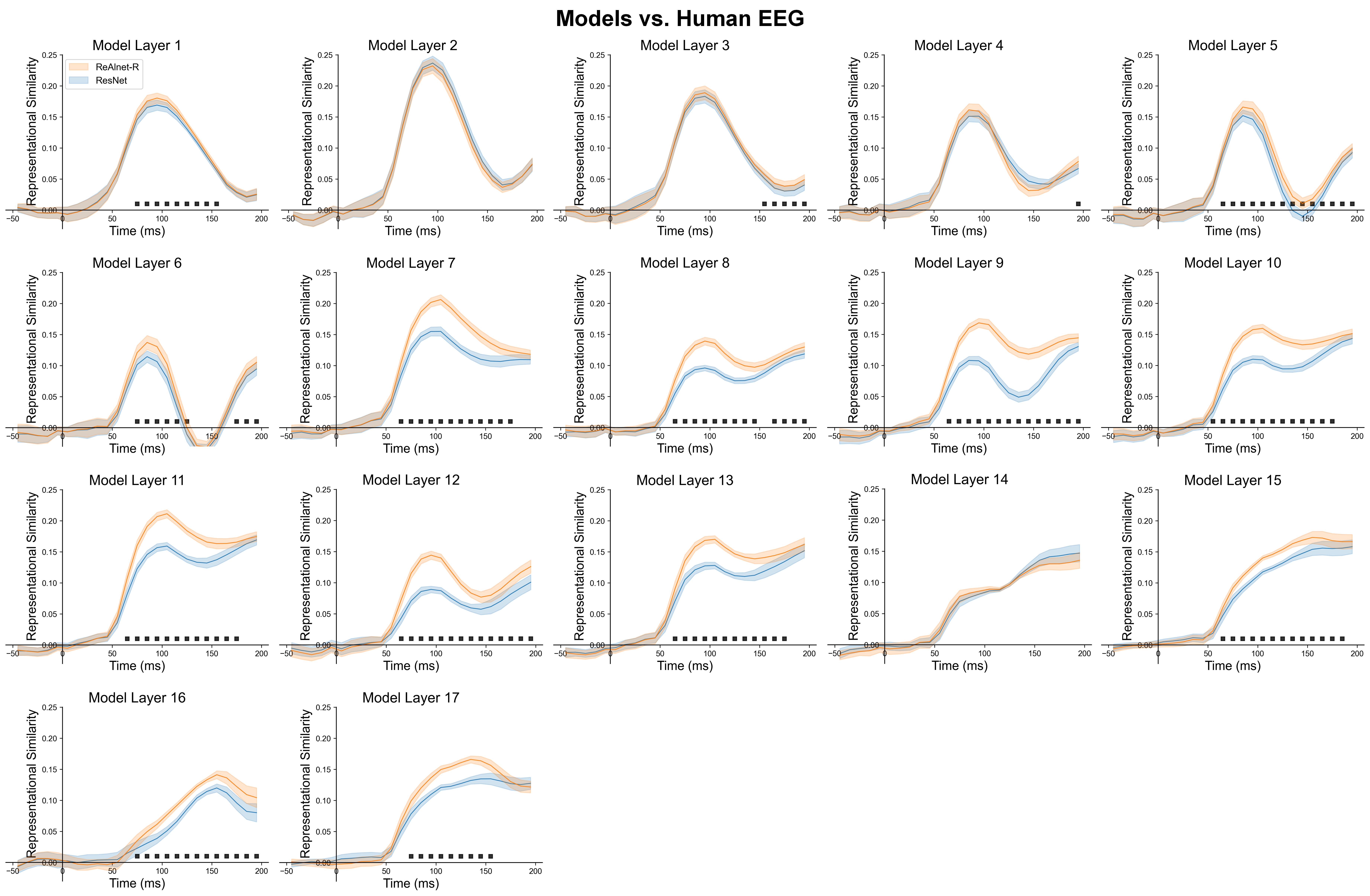}}
\caption{Representational similarity time courses between human EEG and models (ReAlnet-Rs and ResNet) for all 17 layers in ResNet architecture respectively. Black square dots at the bottom indicate the timepoints where ReAlnet-Rs were significantly higher than ResNet ($p<.05$). Lines and shading reflect mean±SEM.}
\label{FigureS8}
\end{center}
\end{figure}

\newpage

\begin{figure}[h!]
\begin{center}
\centerline{\includegraphics[width=\columnwidth]{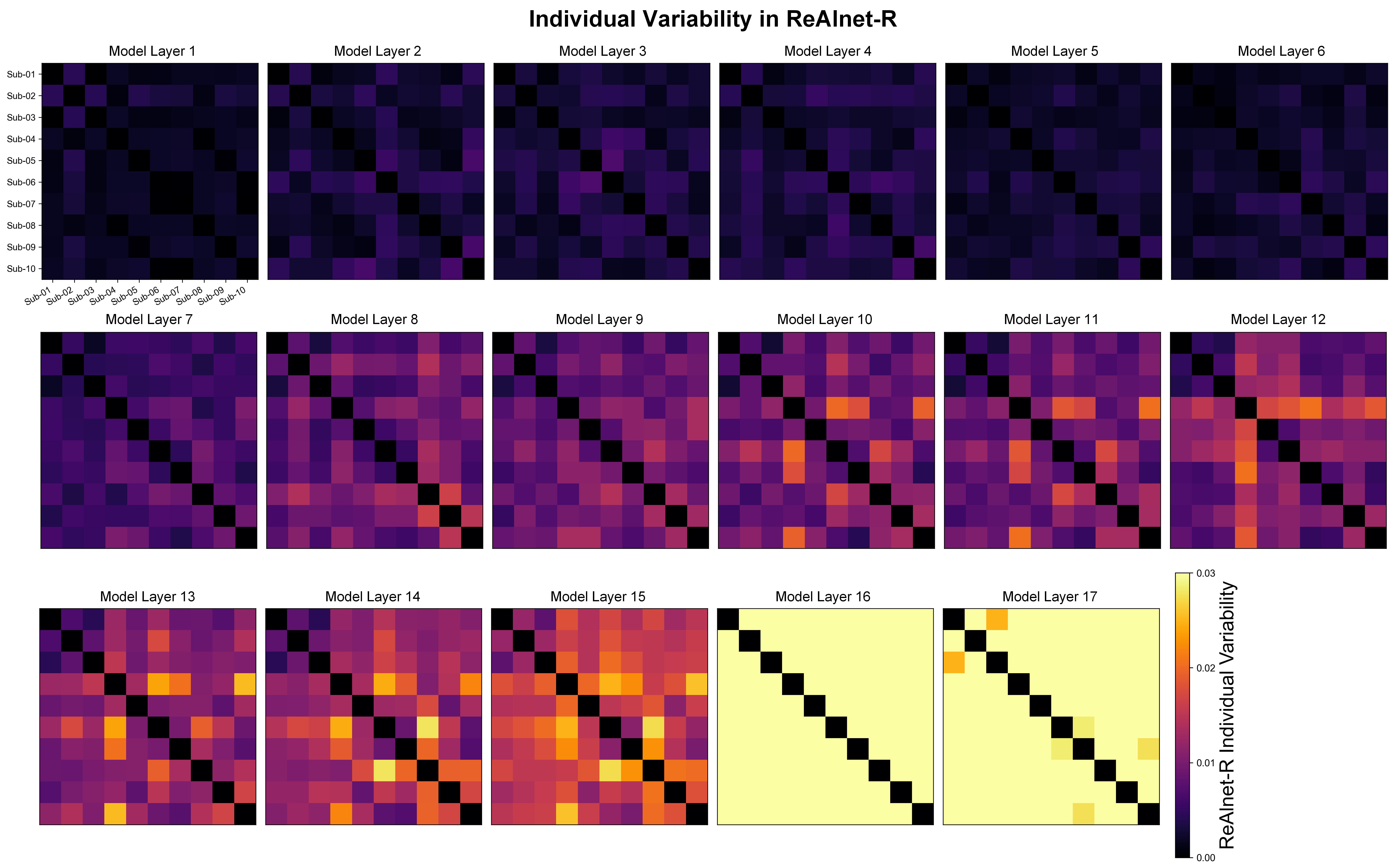}}
\caption{ReAlnet-R individual variability matrices of all layers. Each circle dot indicates a pair of two personalized ReAlnet-Rs.}
\label{FigureS9}
\end{center}
\end{figure}

\newpage

\begin{figure}[h!]
\begin{center}
\centerline{\includegraphics[width=\columnwidth]{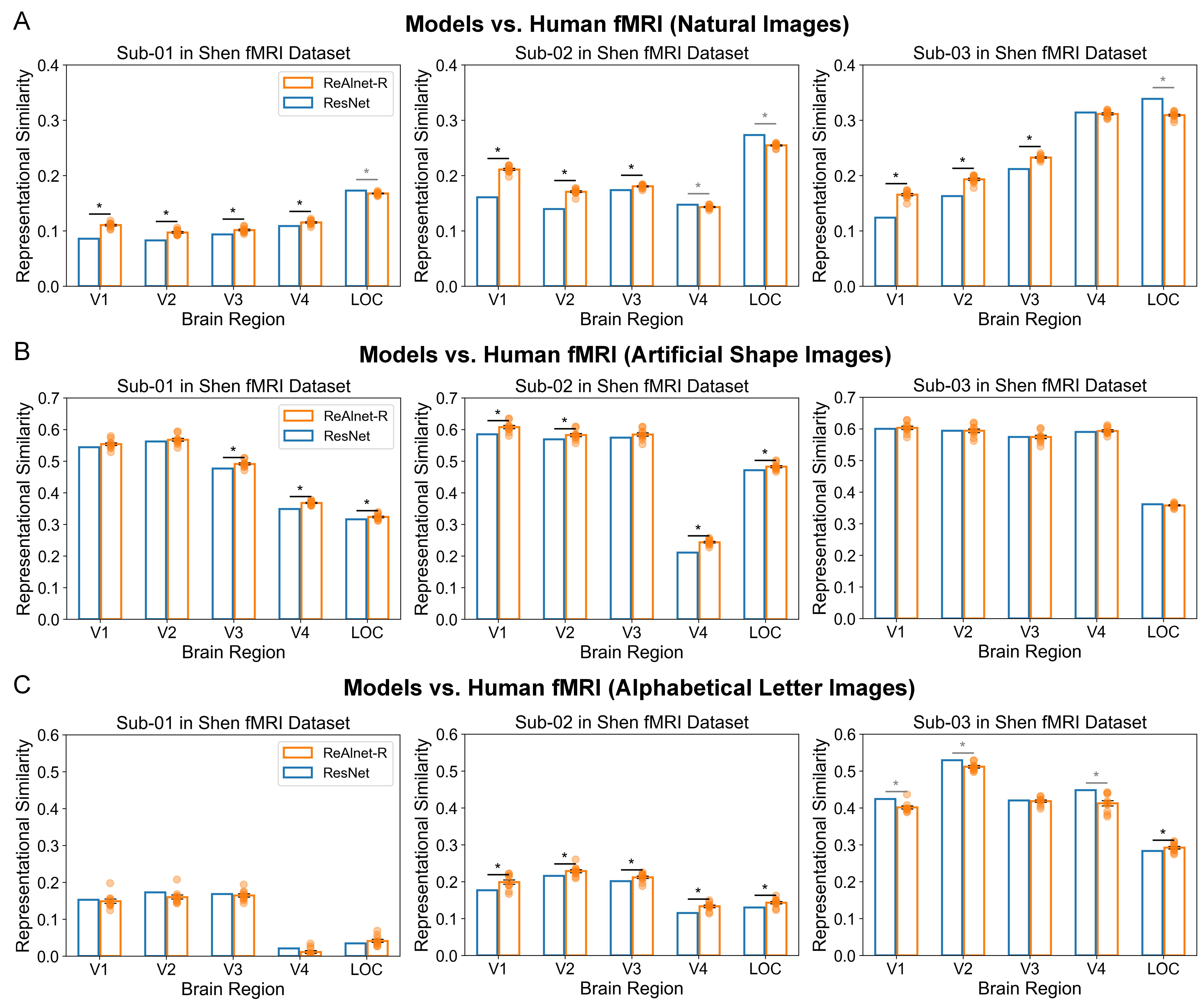}}
\caption{Representational similarity between models (ReAlnet-R and ResNet) and human fMRI of five different brain regions when three subjects viewed (A) natural images, (B) artificial shape images, and (C) alphabetical letter images. Black asterisks indicate significantly higher similarity of ReAlnet-Rs than that of Resnet ($p<.05$). Grey asterisks indicate significantly lower similarity of ReAlnet-R than that of Resnet ($p<.05$). Each circle dot indicates an individual ReAlnet-R. Error bar reflects ±SEM.}
\label{FigureS10}
\end{center}
\end{figure}

\newpage

\begin{figure}[ht]
\begin{center}
\centerline{\includegraphics[width=\columnwidth]{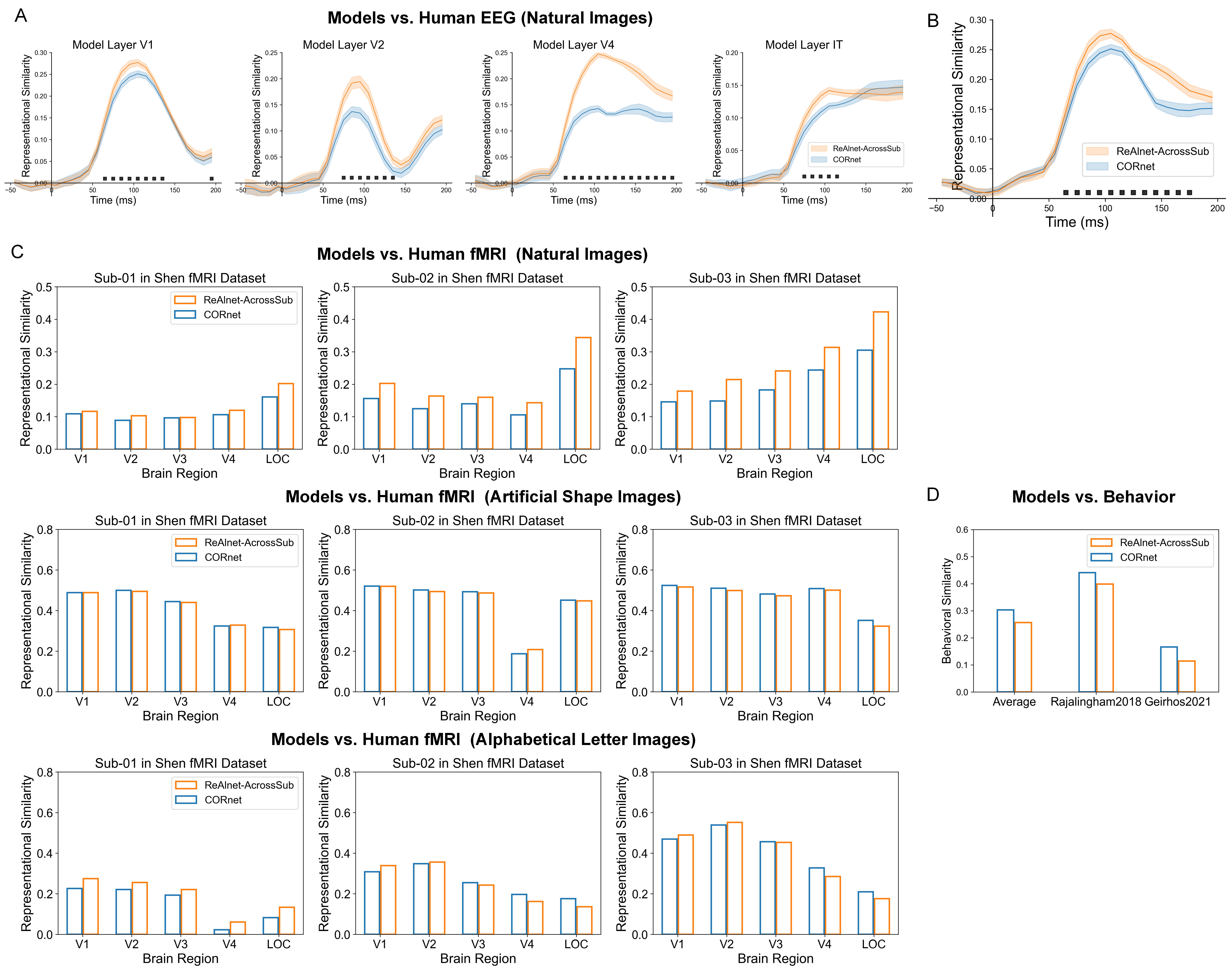}}
\caption{Similar improvements in ReAlnet-AcrossSub. (A) Representational similarity time courses between human EEG and models (ReAlnet-AcrossSub and CORnet) for different layers respectively. Black square dots at the bottom indicate the timepoints where ReAlnet-AcrossSub vs. CORnet were significantly different ($p<.05$). Lines and shading reflect mean±SEM. (B) Time courses of the maximum representational similarity between human EEG and different models (ReAlnet-AcrossSub and CORnet), computed by taking the highest similarity across all model layers at each timepoint. Black square dots at the bottom indicate timepoints where ReAlnet-AcrossSub significantly outperformed CORnet ($p<.05$). Lines and shading reflect mean±SEM. (C) Representational similarity between models and human fMRI of five different brain regions when three subjects in Shen fMRI test dataset viewed natural images, artificial shape images, and alphabetical letter images. (D) Similarity between models and human behavior based on the Brain-Score platform.}
\label{FigureS11}
\end{center}
\end{figure}

\newpage

\begin{figure}[h!]
\begin{center}
\centerline{\includegraphics[width=\columnwidth]{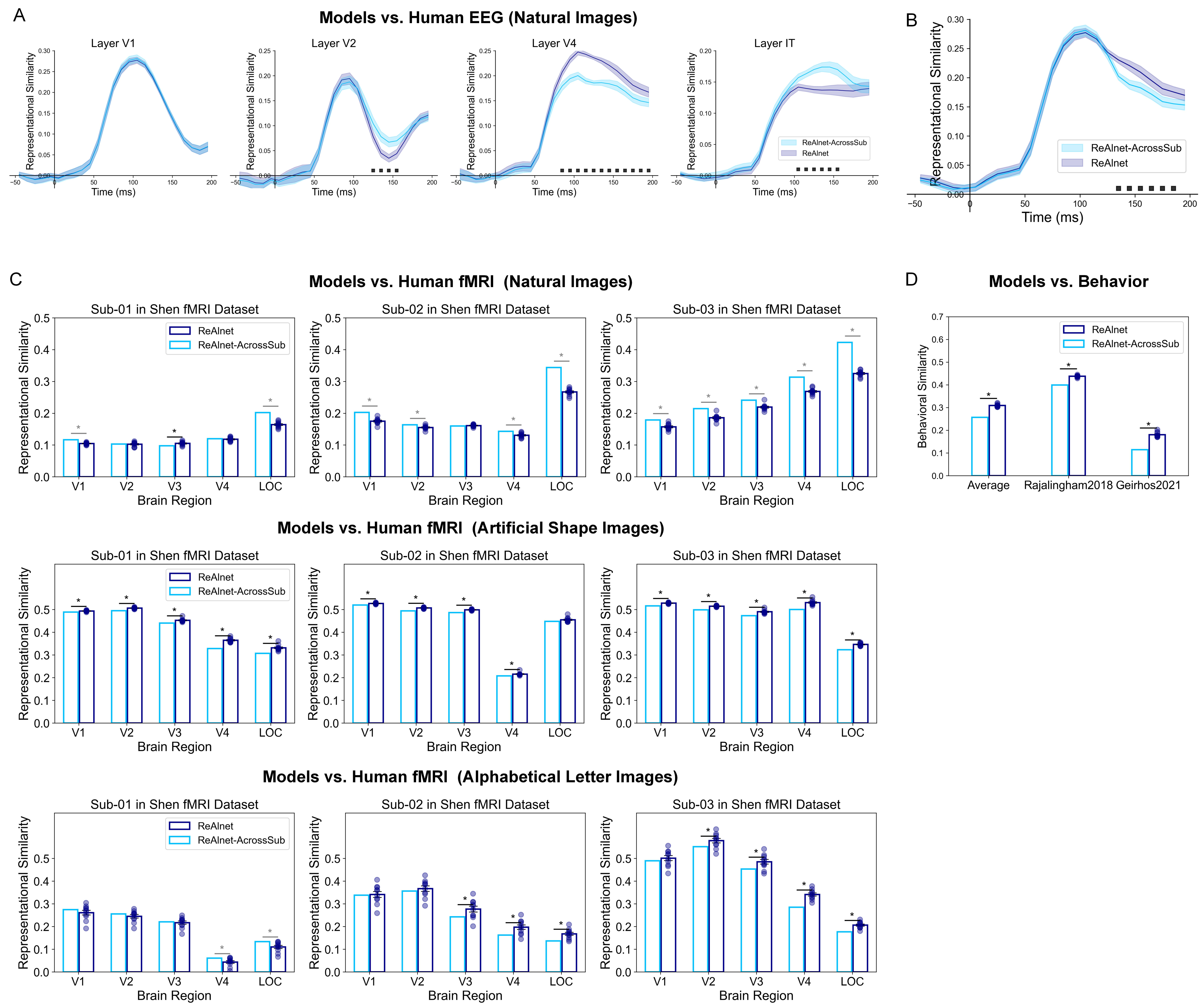}}
\caption{Comparisons between ReAlnet-AcrossSub in ReAlnets. (A) Representational similarity time courses between human EEG and models (ReAlnet-AcrossSub and ReAlnets) for all visual layers. Black square dots at the bottom indicate the timepoints where ReAlnet-AcrossSub vs. ReAlnets were significantly different ($p<.05$). Lines and shading reflect mean±SEM. (B) Time courses of the maximum representational similarity between human EEG and different models (ReAlnet-AcrossSub and ReAlnets), computed by taking the highest similarity across all model layers at each timepoint. Black square dots at the bottom indicate timepoints where ReAlnets significantly outperformed ReAlnet-AcrossSub ($p<.05$). Lines and shading reflect mean±SEM. (D) Representational similarity between models and human fMRI of five different brain regions when three Subjects in Shen fMRI test dataset viewed natural, artificial shape, and alphabetical letter images. Black asterisks indicate significantly higher similarity of ReAlnets than that of ReAlnet-AcrossSub. Grey asterisks indicate significantly lower similarity of ReAlnets than that of ReAlnet-AcrossSub ($p<.05$). Each circle dot indicates an individual ReAlnet. Error bar reflects ±SEM. (E) Similarity between models and human behavior based on the Brain-Score platform. Each circle dot indicates an individual ReAlnet. Error bar reflects ±SEM.}
\label{FigureS12}
\end{center}
\end{figure}

\newpage

\begin{figure}[h!]
\begin{center}
\centerline{\includegraphics[width=\columnwidth]{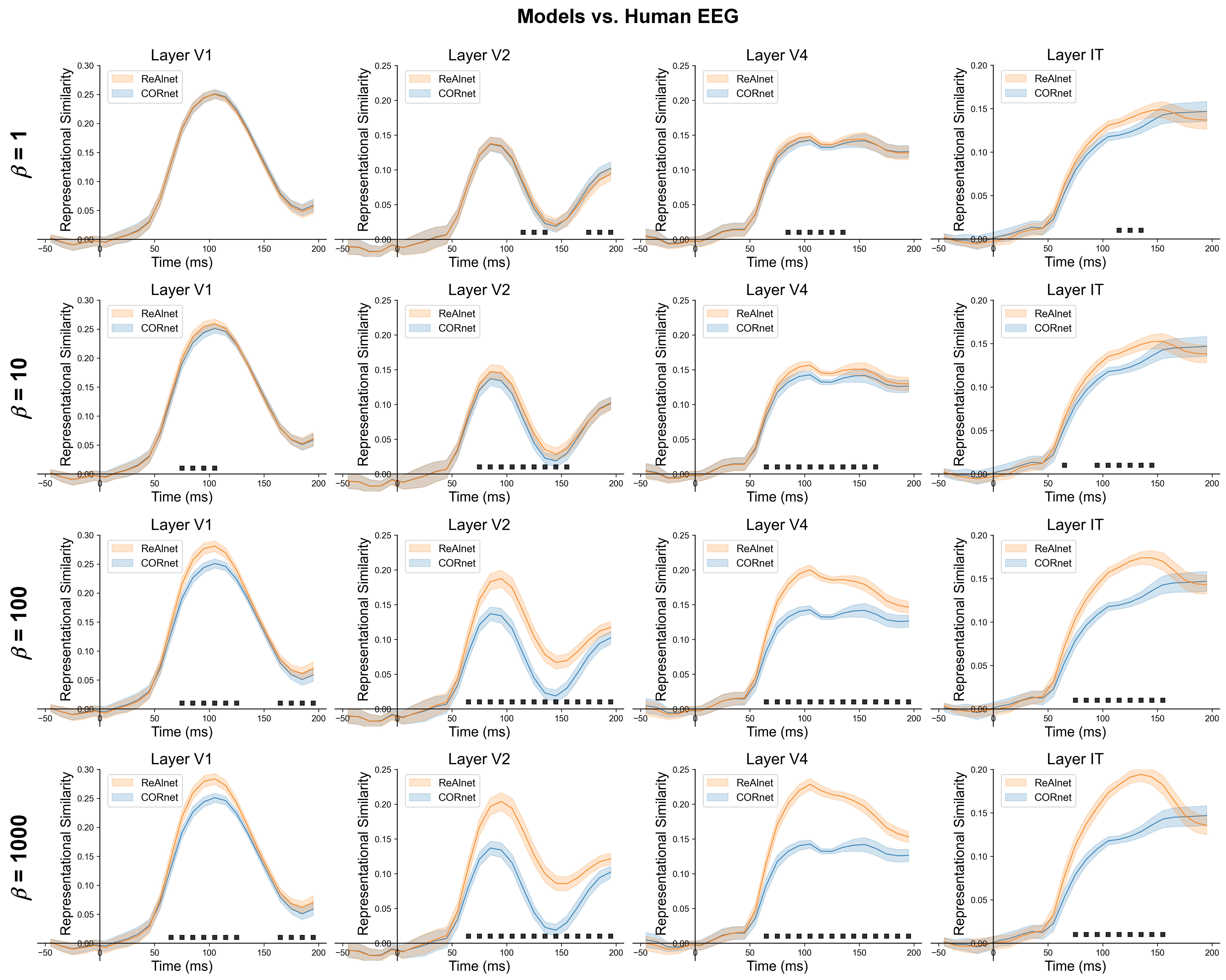}}
\caption{Representational similarity time courses between human EEG and ReAlnets with different loss weights ($\beta$=1, 10, 100, 1000, corresponding to different rows) for different layers respectively. Black square dots at the bottom indicate significant timepoints ($p<.05$). Lines and shading reflect mean ±SEM.}
\label{FigureS13}
\end{center}
\end{figure}

\newpage

\begin{figure}[h!]
\begin{center}
\centerline{\includegraphics[width=\columnwidth]{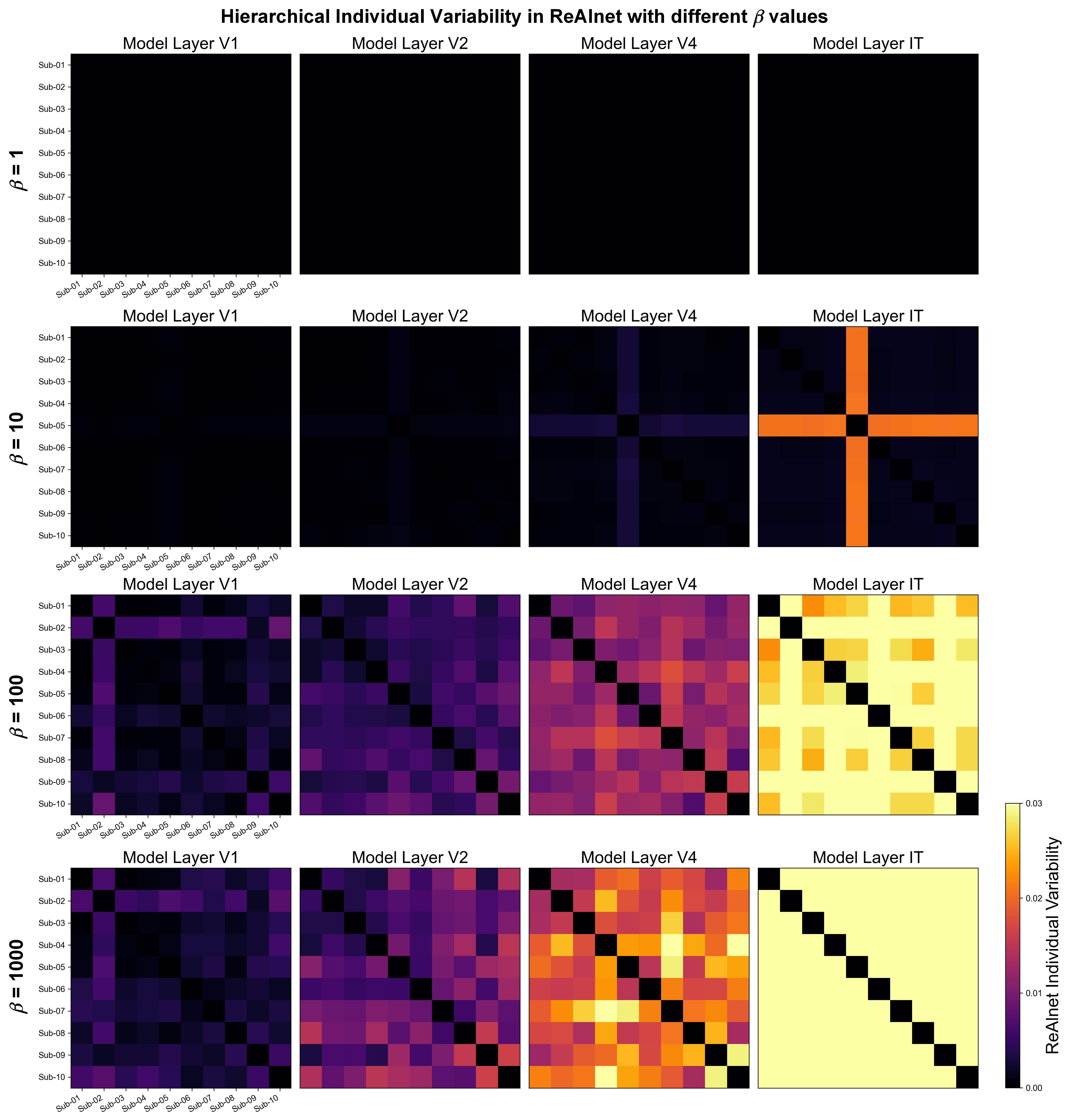}}
\caption{Individual variability matrices of four visual layers of different ReAlnets.}
\label{FigureS14}
\end{center}
\end{figure}

\newpage

\begin{figure}[h!]
\begin{center}
\centerline{\includegraphics[width=\columnwidth]{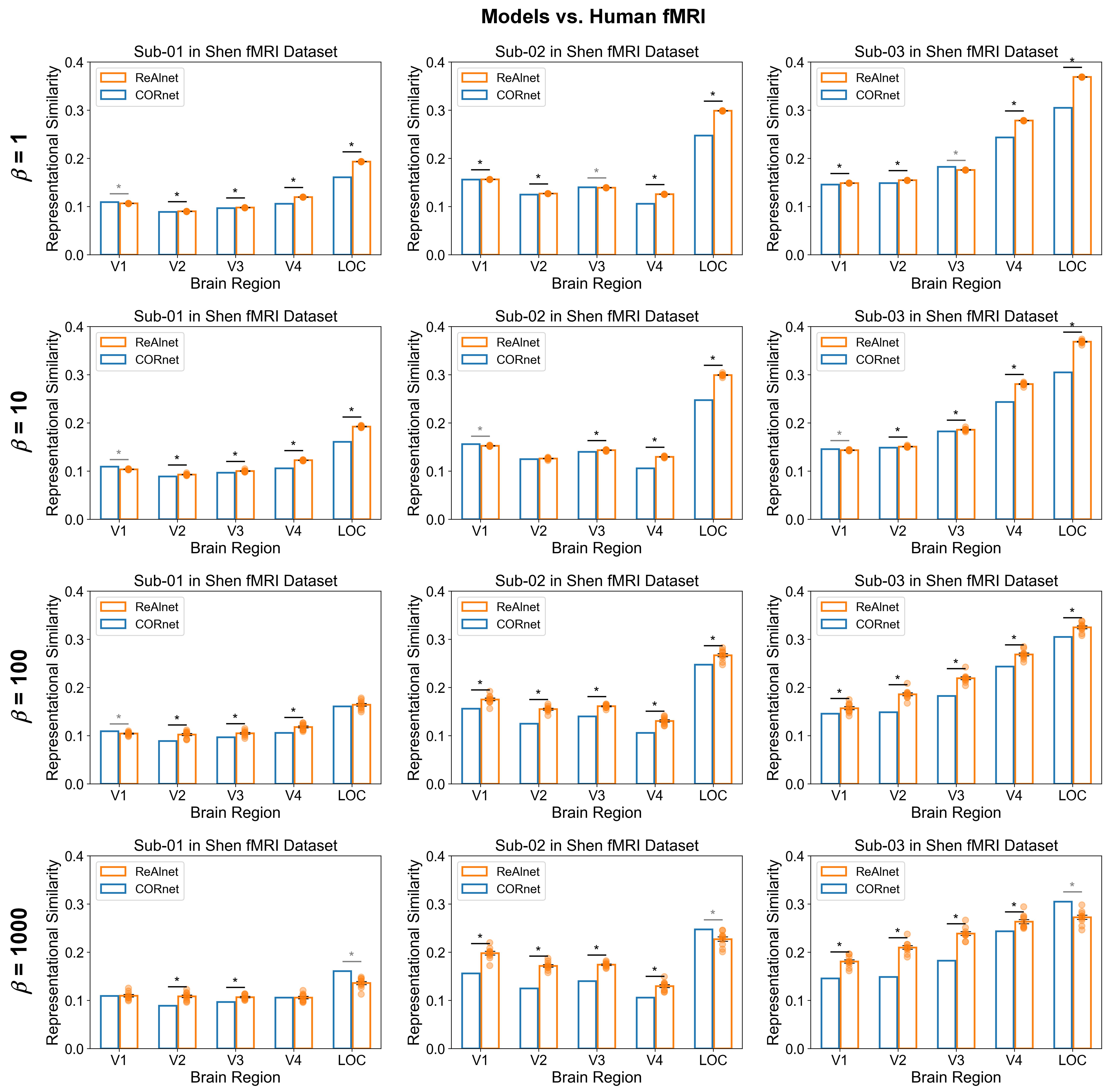}}
\caption{Representational similarity between three subjects’ fMRI activity of five different brain regions when they viewed natural images and different ReAlnets respectively. Black asterisks indicate significantly higher similarity of ReAlnets than that of CORnet ($p<.05$). Grey asterisks indicate significantly higher similarity of ReAlnets than that of CORnet ($p<.05$). Error bar reflects ±SEM.}
\label{FigureS15}
\end{center}
\end{figure}

\newpage

\begin{figure}[h!]
\begin{center}
\centerline{\includegraphics[width=0.6\columnwidth]{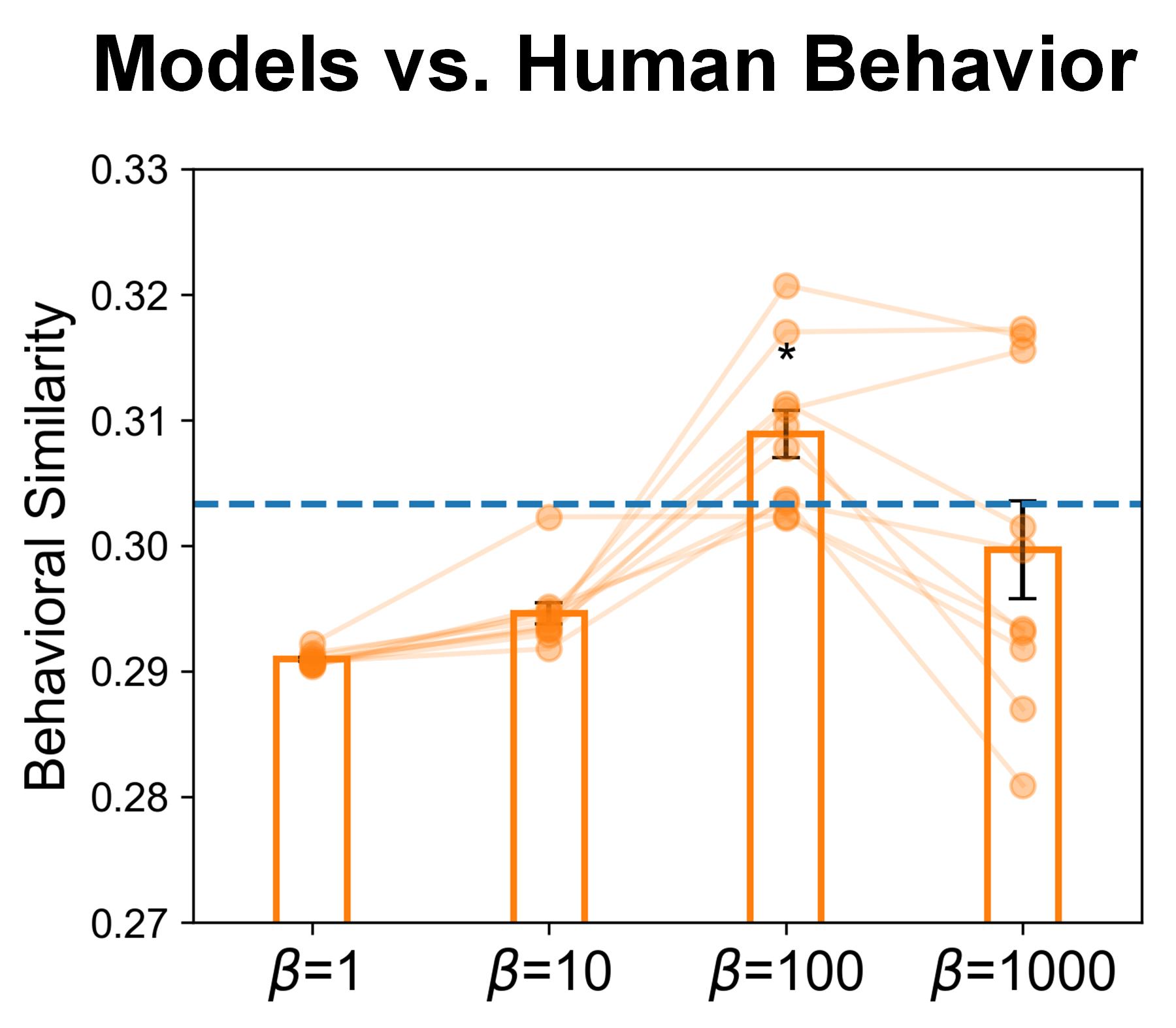}}
\caption{Similarity between human behavior and different ReAlnets based on the Brain-Score platform. Blue dotted line indicates the behavior similarity of CORnet. Each circle dot indicates an individual ReAlnet. Asterisks indicate significantly higher similarity of ReAlnets than that of CORnet ($p<.05$). Error bar reflects ±SEM.}
\label{FigureS16}
\end{center}
\end{figure}

\newpage

\begin{table}[ht]
\begin{center}
\centering
\tiny
\begin{tabular}{|c|c|c|c|}
     \hline
     Comparison & Sub-01 & Sub-02 & Sub-03 \\
     \hline
     \multicolumn{4}{|c|}{Natural Images - Brain Region V1} \\
     \hline
     CORnet vs. ReAlnet & $t=4.1261, p=.0026, d=1.3048$ & \textbf{\boldmath $t=-6.4639, p=.0001, d=-2.0441$} & $t=-3.5726, p=.0060, d=-1.1298$ \\
     \hline
     Scrambled vs. ReAlnet & $t=2.7100, p=.0240, d=1.3129$ & \textbf{\boldmath $t=-6.1716, p=.0002, d=-2.9256$} & $t=-2.2066, p=.0548, d=-1.1493$ \\
     \hline
     \multicolumn{4}{|c|}{Natural Images - Brain Region V2} \\
     \hline
     CORnet vs. ReAlnet & \textbf{\boldmath $t=-5.7036, p=.0003, d=-1.8036$} & \textbf{\boldmath $t=-14.8925, p<.0001, d=-4.7094$} & \textbf{\boldmath $t=-11.3165, p<.0001, d=-3.5786$} \\
     \hline
     Scrambled vs. ReAlnet & \textbf{\boldmath $t=-5.0672, p=.0007, d=-2.8884$} &  \textbf{\boldmath $t=-13.1889, p<.0001, d=-6.3738$} & \textbf{\boldmath $t=-7.4073, p<.0001, d=-3.7776$} \\
     \hline
     \multicolumn{4}{|c|}{Natural Images - Brain Region V3} \\
     \hline
     CORnet vs. ReAlnet & $t=-4.4934, p=.0015, d=-1.4209$ & \textbf{\boldmath $t=-17.3573, p<.0001, d=-5.4888$} & \textbf{\boldmath $t=-11.6277,p <.0001, d=-3.6770$} \\
     \hline
     Scrambled vs. ReAlnet & $t=-3.2830, p=.0095, d=-1.9582$ & \textbf{\boldmath $t=-16.6644, p<.0001, d=-8.5945$} &  \textbf{\boldmath $t=-9.5195, p<.0001, d=-4.8229$} \\
     \hline
     \multicolumn{4}{|c|}{Natural Images - Brain Region V4} \\
     \hline
     CORnet vs. ReAlnet & \textbf{\boldmath $t=-5.6723, p=.0003, d=-1.7937$} & \textbf{\boldmath $t=-10.6293, p<.0001, d=-3.3613$} & \textbf{\boldmath $t=-8.2435, p<.0001, d=-2.6068$} \\
     \hline
     Scrambled vs. ReAlnet & \textbf{\boldmath $t=-6.1886, p=.0002, d=-3.5471$} & \textbf{\boldmath $t=-8.2548, p<.0001, d=-4.3744$} & \textbf{\boldmath $t=-12.2384, p<.0001, d=-6.1740$} \\
     \hline
     \multicolumn{4}{|c|}{Natural Images - Brain Region LOC} \\
     \hline
     CORnet vs. ReAlnet & $t=-1.1089, p=.2962, d=-0.3507$ & \textbf{\boldmath $t=-5.6540, p=.0003, d=-1.7880$} & \textbf{\boldmath $t=-6.3781, p=.0001, d=-2.0169$}\\
     \hline
     Scrambled vs. ReAlnet & \textbf{\boldmath $t=-7.3478, p<.0001, d=-3.3292$} & $t=-4.3087, p=.0020, d=-2.0951$ & $t=-2.6475, p=.0266, d=-1.3932$ \\
     \hline
     \multicolumn{4}{|c|}{Artificial Shape Images - Brain Region V1} \\
     \hline
     CORnet vs. ReAlnet & \textbf{\boldmath $t=-5.4732, p=.0004, d=-1.7308$} & \textbf{\boldmath $t=-7.1874, p<.0001, d=-2.2729$} & \textbf{\boldmath $t=-5.0016, p=.0007, d=-1.5816$} \\
     \hline
     Scrambled vs. ReAlnet & \textbf{\boldmath $t=-5.5769, p=.0003, d=-2.3410$} & \textbf{\boldmath $t=-8.8215, p<.0001, d=-3.7266$} & \textbf{\boldmath $t=-7.1573, p<.0001, d=-3.0877$} \\
     \hline
     \multicolumn{4}{|c|}{Artificial Shape Images - Brain Region V2} \\
     \hline
     CORnet vs. ReAlnet & \textbf{\boldmath $t=-5.5054, p=.0004, d=-1.7410$} & \textbf{\boldmath $t=-5.3422, p=.0005, d=-1.6894$}
     & \textbf{\boldmath $t=-6.8536, p<.0001, d=-2.9450$}  \\
     \hline
     Scrambled vs. ReAlnet & $t=-4.6164, p=.0013, d=-1.4598$ & \textbf{\boldmath $t=-8.1057, p<.0001, d=-3.3251$} & \textbf{\boldmath $t=-6.4433, p=.0001, d=-2.8012$} \\
     \hline
     \multicolumn{4}{|c|}{Artificial Shape Images - Brain Region V3} \\
     \hline
     CORnet vs. ReAlnet & $t=-3.4159, p=.0077, d=-1.0802$ & \textbf{\boldmath $t=-6.8620, p<.0001, d=-2.1700$} & $t=-3.2408, p=.0101, d=-1.0248$ \\
     \hline
     Scrambled vs. ReAlnet & $t=-3.6286, p=.0055, d=-1.6007$ & \textbf{\boldmath $t=-8.4069, p<.0001, d=-3.5363$} & $t=-4.7413, p=.0011, d=-1.8254$ \\
     \hline
     \multicolumn{4}{|c|}{Artificial Shape Images - Brain Region V4} \\
     \hline
     CORnet vs. ReAlnet & \textbf{\boldmath $t=-12.9847, p<.0001, d=-4.1061$} & \textbf{\boldmath $t=-12.9830, p<.0001, d=-4.1056$} & \textbf{\boldmath $t=-6.1701, p=.0002, d=-1.9512$} \\
     \hline
     Scrambled vs. ReAlnet & \textbf{\boldmath $t=-5.5997, p=.0003, d=-2.2670$} & $t=-1.8581, p=.0961, d=0.7624$ & $t=-1.8007, p=.1053, d=-0.6295$ \\
     \hline
     \multicolumn{4}{|c|}{Artificial Shape Images - Brain Region LOC} \\
     \hline
     CORnet vs. ReAlnet & $t=-3.3891, p=.0080, d=-1.0717$ & $t=-0.9881, p=.3489, d=-0.3125$ & $t=2.5743, p=.0300, d=0.8141$ \\
     \hline
     Scrambled vs. ReAlnet & $t=1.4270,p=.1873, d=0.6172$ & $t=3.9425, p=.0034, d=1.3082$ & \textbf{\boldmath $t=6.2139, p=.0002, d=2.4700$} \\
     \hline
     \multicolumn{4}{|c|}{Alphabetical Letter Images - Brain Region V1} \\
     \hline
     CORnet vs. ReAlnet & $t=-3.2981, p=.0093, d=-1.0430$ & $t=-2.4272, p=.0382, d=-0.7676$ & $t=-2.7143, p=.0238, d=-0.8583$ \\
     \hline
     Scrambled vs. ReAlnet & $t=-3.5630, p=.0061, d=-1.6416$ & $t=-1.9204, p=.0870, d=-0.8722$ & $t=-2.1053, p=.0646, d=-0.9906$ \\
     \hline
     \multicolumn{4}{|c|}{Alphabetical Letter Images - Brain Region V2} \\
     \hline
     CORnet vs. ReAlnet & 
     $t=-2.8977, p=.0177, d=-0.9163$ & $t=-1.4172, p=.1901, d=-0.4482$ & $t=-3.8711, p=.0038, d=-1.2242$ \\
     \hline
     Scrambled vs. ReAlnet & $t=-3.8664, p=.0038, d=-1.8001$ & $t=-.8639, p=.4101, d=-0.3880$ & $t=-2.1112, p=.0639, d=-0.9850$ \\
     \hline
     \multicolumn{4}{|c|}{Alphabetical Letter Images - Brain Region V3} \\
     \hline
     CORnet vs. ReAlnet & 
     $t=-3.1275, p=.0122, d=-0.9890$ & $t=-1.6528, p=.1328, d=-0.5227$ & $t=-2.7628, p=.0220, d=-0.8737$ \\
     \hline
     Scrambled vs. ReAlnet & $t=-4.1184, p=.0026, d=-1.9571$ & $t=-.6495, p=.5322, d=-0.3083$ & $t=-1.6782, p=.1276, d=-0.8737$ \\
     \hline
     \multicolumn{4}{|c|}{Alphabetical Letter Images - Brain Region V4} \\
     \hline
     CORnet vs. ReAlnet & $t=-3.4396, p=.0074, d=-1.0877$ & $t=-.0387, p=.9700, d=-0.0122$ & $t=-1.8718, p=.0940, d=-0.5919$ \\
     \hline
     Scrambled vs. ReAlnet & \textbf{\boldmath $t=-5.2222, p=.0006, d=-2.5231$} & $t=.05860, p=.9646, d=0.0279$ & $t=-0.3898, p=.7057, d=-0.1749$ \\
     \hline
     \multicolumn{4}{|c|}{Alphabetical Letter Images - Brain Region LOC} \\
     \hline
     CORnet vs. ReAlnet & 
     $t=-3.9180, p=.0035, d=1.2390$ & $t=1.2020, p=.2600 d=0.3801$ & $t=.6582, p=.5269, d=0.2081$ \\
     \hline
     Scrambled vs. ReAlnet & \textbf{\boldmath $t=-5.0619, p=.0007, d=2.3544$} & $t=1.2307, p=.2496, d=0.5933$ & $t=2.3300, p<.0447, d=1.0685$ \\
     \hline
     
\end{tabular}
\caption{Statistical results, including t-value, p-value, and Cohen's d, of the model-fMRI similarity difference between ReAlnets, Scrambled models, and CORnet, corresponding to \Cref{Figure3}. Bold values indicate particularly strong effects ($p<.001$ and |Cohen's $d$|>1.0).}
\label{TableS1}
\end{center}
\end{table}

\newpage

\begin{table}[ht]
\begin{center}
\centering
\tiny
\begin{tabular}{|c|c|c|c|}
     \hline
     Comparison & Sub-01 & Sub-02 & Sub-03 \\
     \hline
     \multicolumn{4}{|c|}{Brain Region V1} \\
     \hline
     ReAlnet vs. W/o ContLoss &
     $t=-4.7466, p=.0010, d=-1.0280$ &
     $t=.1569, p=.8788, d=.0332$ &
     $t=-4.0000, p=.0031, d=-.5181$ \\
     \hline
     ReAlnet vs. W/o MSELoss &
     \textbf{\boldmath $t=-4.9306, p=.0008, d=-1.7470$} &
     \textbf{\boldmath $t=6.2222, p=.0002, d=2.7874$} &
     $t=1.8787, p=.0930, d=.8756$ \\
     \hline
     ReAlnet vs. Unpaired &
     $t=-2.3628, p=.0424, d=-1.2044$ &
     \textbf{\boldmath $t=6.7134, p<.0001, d=3.1361$} &
     $t=3.2063, p=.0107, d=1.2519$ \\
     \hline
     ReAlnet vs. Scrambled &
     $t=-2.7101, p=.0240, d=-1.3129$ &
     \textbf{\boldmath $t=6.1716, p=.0002, d=2.9256$} &
     $t=2.2066, p=.0548, d=1.1493$ \\
     \hline
     \multicolumn{4}{|c|}{Brain Region V2} \\
     \hline
     ReAlnet vs. W/o ContLoss &
     $t=.5237, p=.6132, d=.1676$ &
     $t=.8922, p=.3955, d=.2297$ &
     $t=-.5701, p=.5825, d=-.1240$ \\
     \hline
     ReAlnet vs. W/o MSELoss &
     \textbf{\boldmath $t=7.3831, p<.0001, d=3.4361$} &
     \textbf{\boldmath $t=12.7512, p<.0001, d=6.1594$} &
     \textbf{\boldmath $t=9.0401, p<.0001, d=3.9936$} \\
     \hline
     ReAlnet vs. Unpaired &
     $t=3.5303, p=.0064, d=1.4620$ &
     \textbf{\boldmath $t=12.3683, p<.0001, d=6.2417$} &
     \textbf{\boldmath $t=7.0155, p<.0001, d=3.1526$} \\
     \hline
     ReAlnet vs. Scrambled &
     \textbf{\boldmath $t=5.0672, p=.0007, d=2.8884$} &
     \textbf{\boldmath $t=13.1889, p<.0001, d=6.3738$} &
     \textbf{\boldmath $t=7.4073 p<.0001, d=3.7776$} \\
     \hline
     \multicolumn{4}{|c|}{Brain Region V3} \\
     \hline
     ReAlnet vs. W/o ContLoss &
     $t=.4579, p=.6579, d=.1595$ &
     $t=.6156, p=.5534, d=.1915$ &
     $t=-.6427, p=.5364, d=-.1111$ \\
     \hline
     ReAlnet vs. W/o MSELoss &
     \textbf{\boldmath $t=5.3837, p=.0004, d=2.7160$} &
     \textbf{\boldmath $t=29.0426, p<.0001, d=11.9132$} & \textbf{\boldmath $t=12.5099, p<.0001, d=5.7474$} \\
     \hline
     ReAlnet vs. Unpaired &
     $t=4.0800, p=.0028, d=1.5770$ &
     \textbf{\boldmath $t=20.9276, p<.0001, d=6.7735$} &
     \textbf{\boldmath $t=9.4683, p<.0001, d=3.8598$} \\
     \hline
     ReAlnet vs. Scrambled &
     $t=3.2830, p=.0095, d=1.9582$ &
     \textbf{\boldmath $t=16.6645, p<.0001, d=8.5945$} &
     \textbf{\boldmath $t=9.5195, p<.0001, d=4.8229$} \\
     \hline
     \multicolumn{4}{|c|}{Brain Region V4} \\
     \hline
     ReAlnet vs. W/o ContLoss &
     $t=4.5455, p=.0014, d=1.4354$ &
     \textbf{\boldmath $t=8.2963, p<.0001, d=1.4072$} &
     \textbf{\boldmath $t=6.3084, p=.0001, d=1.6754$} \\
     \hline
     ReAlnet vs. W/o MSELoss &
     \textbf{\boldmath $t=7.9493, p<.0001, d=4.0403$} &
     \textbf{\boldmath $t=12.6584, p<.0001, d=5.8177$} & 
     \textbf{\boldmath $t=14.4217, p<.0001, d=6.6589$} \\
     \hline
     ReAlnet vs. Unpaired &
     \textbf{\boldmath $t=5.8403, p=.0002, d=2.3359$} &
     \textbf{\boldmath $t=11.8323, p<.0001, d=4.1429$} & 
     \textbf{\boldmath $t=10.9515, p<.0001, d=3.2523$} \\
     \hline
     ReAlnet vs. Scrambled &
     \textbf{\boldmath $t=6.1886, p=.0002, d=3.5471$} &
     \textbf{\boldmath $t=8.2545, p<.0001, d=4.3744$} &
     \textbf{\boldmath $t=12.2383, p<.0001, d=6.1740$} \\
     \hline
     \multicolumn{4}{|c|}{Brain Region LOC} \\
     \hline
     ReAlnet vs. W/o ContLoss &
     \textbf{\boldmath $t=16.3594, p<.0001, d=2.3304$} &
     \textbf{\boldmath $t=14.7669, p<.0001, d=3.7446$} &
     \textbf{\boldmath $t=14.5849, p<.0001, d=3.7883$} \\
     \hline
     ReAlnet vs. W/o MSELoss &
     \textbf{\boldmath $t=6.1111, p=.0002, d=2.1028$} &
     \textbf{\boldmath $t=4.9615, p=.0007, d=2.4206$} &
     $t=3.5755, p=.0060, d=1.7802$ \\
     \hline
     ReAlnet vs. Unpaired &
     $t=4.7542, p=.0010, d=2.5211$ &
     \textbf{\boldmath $t=9.4013, p<.0001, d=4.4076$} &
     \textbf{\boldmath $t=7.9380, p<.0001, d=3.7081$} \\
     \hline
     ReAlnet vs. Scrambled &
     \textbf{\boldmath $t=7.3478, p<.0001, d=3.3292$} &
     $t=4.3087, p=.0020, d=2.0951$ &
     $t=2.6475, p=.0266, d=1.3932$ \\
     \hline
\end{tabular}
\caption{Statistical results, including t-value, p-value, and Cohen's d, of the model-fMRI similarity difference between different control models. Bold values indicate particularly strong effects ($p<.001$ and |Cohen's $d$|>1.0).}
\label{TableS2}
\end{center}
\end{table}

\newpage

\begin{table}[h!]
\begin{center}
\centering
\tiny
\begin{tabular}{|c|c|c|}
     \hline
     Comparison  & Human EEG (Within-Modality) & Human fMRI (Cross-Modality) \\
     \hline
     ReAlnet vs. CORnet &
     \textbf{\boldmath $t=12.0891, p<.0001, d=3.8229$} & 
     \textbf{\boldmath $t=11.5082, p<.0001, d=3.6392$} \\
     \hline
     W/o ContLoss vs. CORnet & 
     \textbf{\boldmath $t=11.4987, p<.0001, d=3.6362$} & 
     $t=4.1036, p=.0027, d=1.2977$ \\
     \hline
     W/o MSELoss vs. CORnet &
     $t=.5640, p=.5865, d=0.1784$ & 
     \textbf{\boldmath $t=-6.3837, p=.0001, d=-2.0187$} \\
     \hline
     Unpaired vs. CORnet &
     $t=-.6654, p=.5225, d=-.2104$ & 
     $t=-3.1304, p=.0121, d=-0.9900$ \\
     \hline
     Scrambled vs. CORnet &
     $t=1.9450, p=.0835, d=.6154$ & 
     $t=-3.2647, p=.0098, d=-1.0324$ \\
     \hline
     ReAlnet vs. W/o ContLoss & 
     $t=-.8543, p=.4151, d=.0517$ & 
     \textbf{\boldmath $t=8.0364, p<.0001, d=1.5494$} \\
     \hline
     ReAlnet vs. W/o MSELoss & 
     \textbf{\boldmath $t=14.5308, p<.0001, d=5.0884$} & 
     \textbf{\boldmath $t=11.5870, p<.0001, d=5.8832$} \\
     \hline
     ReAlnet vs. Unpaired &
     \textbf{\boldmath $t=14.0731, p<.0001, d=5.2863$} & 
     \textbf{\boldmath $t=13.1332, p<.0001, d=5.2232$} \\
     \hline
     ReAlnet vs. Scrambled &
     \textbf{\boldmath $t=13.3893, p<.0001, d=4.9035$} & 
     \textbf{\boldmath $t=8.1128, p<.0001, d=4.8976$} \\
     \hline
\end{tabular}
\caption{Statistical results, including t-value, p-value, and Cohen's d, of the model-brain (EEG and fMRI) similarity difference between ReAlnets, control models, and CORnet, corresponding to \Cref{Figure5}. Bold values indicate particularly strong effects ($p<.001$ and |Cohen's $d$|>1.0).}
\label{TableS3}
\end{center}
\end{table}

\newpage

\begin{table}[ht]
\begin{center}
\centering
\tiny
\begin{tabular}{|c|c|c|c|}
     \hline
     Comparison & Sub-01 & Sub-02 & Sub-03 \\
     \hline
     \multicolumn{4}{|c|}{Natural Images - Brain Region V1} \\
     \hline
     ResNet vs. ReAlnet-R &
     \textbf{\boldmath $t=-15.7710, p<.0001, d=-4.9872$} &
     \textbf{\boldmath $t=-24.5242, p<.0001, d=-7.7552$} & 
     \textbf{\boldmath $t=-18.0627, p<.0001, d=-5.7119$} \\
     \hline
     \multicolumn{4}{|c|}{Natural Images - Brain Region V2} \\
     \hline
     ResNet vs. ReAlnet-R &
     \textbf{\boldmath $t=-9.1640, p<.0001, d=-2.8978$} &
     \textbf{\boldmath $t=-18.1327, p<.0001, d=-5.7340$} & 
     \textbf{\boldmath $t=-13.0616, p<.0001, d=-4.1304$} \\
     \hline
     \multicolumn{4}{|c|}{Natural Images - Brain Region V3} \\
     \hline
     ResNet vs. ReAlnet-R &
     \textbf{\boldmath $t=-5.1613, p=.0006, d=-1.6321$} &
     \textbf{\boldmath $t=-6.1212, p=.0002, d=-1.9357$} & 
     \textbf{\boldmath $t=-14.2771, p<.0001, d=-4.5148$} \\
     \hline
     \multicolumn{4}{|c|}{Natural Images - Brain Region V4} \\
     \hline
     ResNet vs. ReAlnet-R &
     $t=-4.4650, p=.0016, d=-1.4119$ &
     $t=3.6102, p=.0057, d=1.1416$ & 
     $t=1.3162, p=.2206, d=.4162$ \\
     \hline
     \multicolumn{4}{|c|}{Natural Images - Brain Region LOC} \\
     \hline
     ResNet vs. ReAlnet-R &
     \textbf{\boldmath $t=5.4441, p=.0004, d=1.7216$} &
     \textbf{\boldmath $t=15.7909, p<.0001, d=4.9935$} & 
     \textbf{\boldmath $t=14.8503, p<.0001, d=4.6961$} \\
     \hline
     \multicolumn{4}{|c|}{Artificial Shape Images - Brain Region V1} \\
     \hline
     ResNet vs. ReAlnet-R &
     $t=-2.0918, p=.0660, d=-.6615$ &
     $t=-4.2623, p=.0021, d=-1.3479$ & 
     $t=-.5449, p=.5991, d=-.1723$ \\
     \hline
     \multicolumn{4}{|c|}{Artificial Shape Images - Brain Region V2} \\
     \hline
     ResNet vs. ReAlnet-R &
     $t=-1.1047, p=.2979, d=-.3493$ &
     $t=-2.6039, p=.0286, d=-.8234$ & 
     $t=.1015, p=.9214, d=.0321$ \\
     \hline
     \multicolumn{4}{|c|}{Artificial Shape Images - Brain Region V3} \\
     \hline
     ResNet vs. ReAlnet-R &
     \textbf{\boldmath $t=-3.8858, p=.0037, d=-1.2288$} &
     \textbf{\boldmath $t=-6.4639, p=.0001, d=-2.0441$} & 
     $t=.0731, p=.9433, d=.0231$ \\
     \hline
     \multicolumn{4}{|c|}{Artificial Shape Images - Brain Region V4} \\
     \hline
     ResNet vs. ReAlnet-R &
     \textbf{\boldmath $t=-9.1432, p<.0001, d=-2.8913$} &
     \textbf{\boldmath $t=-11.2972, p<.0001, d=-3.5725$} & 
     $t=-.9364, p=.3735, d=-.2961$ \\
     \hline
     \multicolumn{4}{|c|}{Artificial Shape Images - Brain Region LOC} \\
     \hline
     ResNet vs. ReAlnet-R &
     $t=-2.4581, p=.0363, d=-.7773$ &
     $t=-3.1366, p=.0120, d=-.9919$ & 
     $t=1.4669, p=.1764, d=.4639$ \\
     \hline
     \multicolumn{4}{|c|}{Alphabetical Letter Images - Brain Region V1} \\
     \hline
     ResNet vs. ReAlnet-R &
     $t=.6018, p=.5621, d=.1903$ &
     $t=-3.5423, p=.0063, d=-1.1202$ & 
     \textbf{\boldmath $t=5.2424, p=.0005, d=1.6578$} \\
     \hline
     \multicolumn{4}{|c|}{Alphabetical Letter Images - Brain Region V2} \\
     \hline
     ResNet vs. ReAlnet-R &
     $t=2.1266, p=.0624, d=.6725$ &
     $t=-2.7690, p=.0218, d=-.8756$ & 
     \textbf{\boldmath $t=5.1219, p=.0006, d=1.6197$} \\
     \hline
     \multicolumn{4}{|c|}{Alphabetical Letter Images - Brain Region V3} \\
     \hline
     ResNet vs. ReAlnet-R &
     $t=.8997, p=.0026, d=.2845$ &
     $t=-2.7168, p=.0237, d=-.8591$ & 
     $t=.5619, p=.5879, d=.1178$ \\
     \hline
     \multicolumn{4}{|c|}{Alphabetical Letter Images - Brain Region V4} \\
     \hline
     ResNet vs. ReAlnet-R &
     $t=2.5011, p=.0338, d=.7909$ &
     \textbf{\boldmath $t=-5.0546, p=.0007, d=-1.5984$} & 
     $t=4.6986, p=.0011, d=1.4858$ \\
     \hline
     \multicolumn{4}{|c|}{Alphabetical Letter Images - Brain Region LOC} \\
     \hline
     ResNet vs. ReAlnet-R &
     $t=-1.5012, p=.1675, d=.4747$ &
     $t=-3.5147, p=.0066, d=-1.1114$ & 
     $t=-2.4022, p=.0398, d=.7597$ \\
     \hline
     
\end{tabular}
\caption{Statistical results, including t-value, p-value, and Cohen's d, of the model-fMRI similarity difference between ReAlnet-Rs and ResNet, corresponding to \Cref{Figure6}C. Bold values indicate particularly strong effects ($p<.001$ and |Cohen's $d$|>1.0).}
\label{TableS4}
\end{center}
\end{table}

\end{document}